\pdfoutput=1 
\PassOptionsToPackage{table}{xcolor}
\documentclass[sigconf,authorversion,nonacm]{acmart}
\AtBeginDocument{%
  \providecommand\BibTeX{{%
    \normalfont B\kern-0.5em{\scshape i\kern-0.25em b}\kern-0.8em\TeX}}}

\acmConference[KDD ADS'23]{29th ACM SIGKDD Conference on Knowledge Discovery and Data Mining}{August 06--15,
  2023}{Long Beach, CA}
\acmPrice{15.00}
\acmISBN{978-1-4503-XXXX-X/18/06}

\usepackage[table]{xcolor}
\usepackage{comment}
\usepackage{booktabs} % For formal tables
\usepackage{multirow}

\usepackage{graphicx}
\graphicspath{ {resource/figure/} }

\setcitestyle{square}

\usepackage{graphicx}
\usepackage[utf8]{inputenc}
\usepackage[misc,geometry]{ifsym} 
\usepackage{color}
\usepackage{hyperref} 
\usepackage[bottom]{footmisc}
\usepackage{supertabular}
\usepackage{afterpage}
\usepackage{url}
\usepackage{multicol}
\usepackage{multirow}

\usepackage{times}
\usepackage{latexsym}
\usepackage{tabularx}
\usepackage{multirow}
\usepackage{booktabs}
\usepackage{color,graphicx} % incluir figuras
\usepackage{colortbl}
\usepackage{subcaption}
\usepackage{multirow}
\usepackage{multicol}
\usepackage{amsmath}
\usepackage[linesnumbered,ruled,vlined]{algorithm2e}

\definecolor{orcidlogo}{rgb}{0.37,0.48,0.13}
\definecolor{unilogo}{rgb}{0.16, 0.26, 0.58}
\definecolor{maillogo}{rgb}{0.58, 0.16, 0.26}
\definecolor{darkblue}{rgb}{0.0,0.0,0.0}
\hypersetup{colorlinks,breaklinks,
            linkcolor=darkblue,urlcolor=darkblue,
            anchorcolor=darkblue,citecolor=darkblue}

\usepackage{mathtools}

\makeatletter
\def\@copyrightspace{\relax}
\makeatother

\begin{document}

\title{A Novel Two-Step Fine-Tuning Pipeline for Cold-Start Active Learning in Text Classification Tasks}

\author{Fabiano Belém}
\affiliation{%
  \institution{Federal University of Minas Gerais}
  \country{Brazil}
}
\email{fmuniz@dcc.ufmg.br}

\author{Washington Cunha}
\affiliation{%
  \institution{Federal University of Minas Gerais}
  \country{Brazil}
}
\email{washingtoncunha@dcc.ufmg.br}

\author{Celso França}
\affiliation{%
  \institution{Federal University of Minas Gerais}
  \country{Brazil}
}
\email{celsofranca@dcc.ufmg.br}

\author{Claudio Andrade}
\affiliation{%
  \institution{Federal University of Minas Gerais}
  \country{Brazil}
}
\email{claudio.valiense@dcc.ufmg.br}

\author{Leonardo Rocha}
\affiliation{%
    \institution{Federal University of São João del-Rei}
    \country{Brazil}
}
\email{lcrocha@ufsj.edu.br}

\author{Marcos André Gonçalves}
\affiliation{%
  \institution{Federal University of Minas Gerais}
  \country{Brazil}
}
\email{mgoncalv@dcc.ufmg.br}

\renewcommand{\shortauthors}{Fabiano Belém, Washington Cunha, Celso França, Claudio Andrade, Leonardo Rocha, \& Marcos André Gonçalves}
\newcommand{\red}[1]{\textcolor{red}{#1}}

\begin{abstract}

This is the first work to investigate the effectiveness of BERT-based contextual embeddings in active learning (AL) tasks on cold-start scenarios, where traditional fine-tuning is infeasible due to the absence of labeled data. Our primary contribution is the proposal of a more robust fine-tuning pipeline - DoTCAL -  that diminishes the reliance on labeled data in AL using two steps: (1) fully leveraging unlabeled data through domain adaptation of the embeddings via masked language modeling and (2) further adjusting model weights using labeled data selected by AL. Our evaluation contrasts BERT-based embeddings with other prevalent text representation paradigms, including Bag of Words (BoW), Latent Semantic Indexing (LSI), and FastText, at two critical stages of the AL process: instance selection and classification. Experiments conducted on eight ATC benchmarks with varying AL budgets (number of labeled instances) and number of instances (about 5,000 to 300,000) demonstrates  DoTCAL's superior effectiveness, achieving up to a 33\% improvement in Macro-F1, while reducing labeling efforts by half compared to the traditional one-step method. We also found that in several tasks, BoW and LSI (due to information aggregation) produce results superior (up to 59\% ) to BERT, especially in low-budget scenarios and hard-to-classify tasks, which is quite surprising.
\end{abstract}


\maketitle

\section{Introduction} \label{sec:intro}

Two of the most challenging issues of the supervised automatic text classification (ATC) task are: (i) how to represent textual data in meaningful, machine-tractable ways~\citep{Cunha_2021} and (ii) how to reduce the costs of obtaining labeled data to train classification models~\citep{Settles2010}. 
Regarding the first issue, contextual embeddings such as those produced by Bidirectional Encoder from Transformers or BERT~\citep{devlin2019bert} have produced state-of-the-art results  for a large (but not exhaustive) range of ATC tasks. These embeddings are pre-trained to capture general linguistic patterns from large external collections of unlabeled documents and can be adjusted to more specific domains by means of a fine-tuning process.\looseness=-1

Solutions to the second issue frequently involve active learning (AL) strategies, which aim to select, among the often abundant set of unlabeled data, only the most ``informative'' (diversified and representative) data instances to label.  An AL process can be divided into two stages, illustrated at the top of Figure 1: (i) the selection stage, which aims at selecting informative instances to label, and (ii) the classification stage, in which we train a classification model using the instances selected in stage (i), after being actively labeled. In each stage, textual data must be properly represented (e.g., with contextual embeddings), illustrated in Fig. 1 as rounded rectangles. At the start of stage (i), no fine-tuning for ATC can be performed since there is still no labeled data, a scenario referred to as \textbf{cold-start}. Besides that, the number of instances selected in stage (i) is typically limited to a given \textit{budget} or the \textit{maximum number of instances} one can afford to label, as the cost and complexity of labeling data are non-negligible.\looseness=-1

\begin{figure*}[t]
\hspace{-0.3cm}
    \centering
    \includegraphics[width=0.925\textwidth]{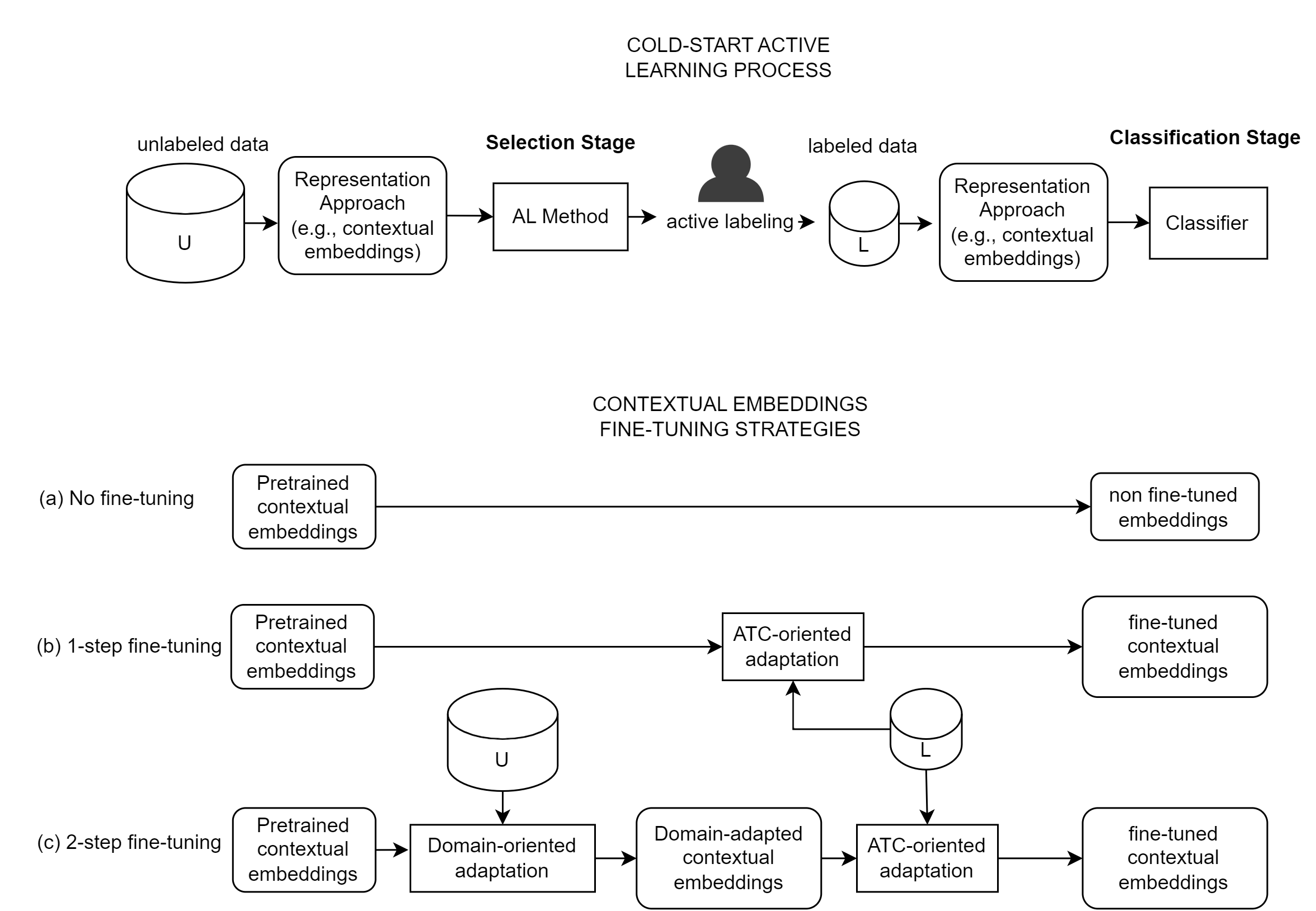}
    \caption{Cold-Start Active Learning for ATC and contextual embeddings fine-tuning approaches}
    \label{fig:al}
\end{figure*}

In this context, we aim to tackle the ``cold-start'' active learning problem for ATC under the (text) representation perspective, that is, we aim to investigate how different representation approaches impact AL effectiveness considering different budget sizes. Despite the success of contextual embeddings for ATC, little has been investigated regarding their  effectiveness in the selection phase of the AL, \textit{especially in a cold-start scenario in which there is no labeled data available (and no fine-tuning is possible)}. One of the few studies that evaluate contextual embeddings for AL~\citep{emnlp_2020} only performs a direct fine-tuning process with already labeled data to adapt the produced embeddings to the target classification task. We refer to this traditional fine-tuning method as \textbf{one-step fine-tuning}, illustrated in Figure 1-b. In most AL settings, the labeled dataset is still under construction and is typically small (due to low budgets in the beginning of the process), which may be detrimental to the effectiveness of this traditional fine-tuning method.\looseness=-1

Thus, it is desirable to invest in more robust fine-tuning processes for AL which reduce the need for labeled data. Accordingly, in this article, as our \textbf{first contribution} we propose to improve AL methods that exploit contextual embeddings by means of a novel \textbf{two-step} \textit{fine-tuning pipeline} named DoTCAL (\textbf{Do}main and \textbf{T}ask Adaptation for \textbf{C}old-Start \textbf{A}ctive \textbf{L}earning (Fig. 1-a), consisting of two phases: (i) a domain-oriented adaptation of the model, exploiting all available unlabeled training data and (ii) a further adaptation of the model using the training samples that were actively labeled after being selected by the AL process (task adaptation). More specifically, while in step (i) we continue the model adaptation to the target domain by pretraining using a masked language modeling (MLM) objective (slot-filling with words), in step (ii), model weights are adjusted using the ATC objective function.  We hypothesize that this new  two-step process allows higher classification effectiveness with a lower labeling effort when compared to the direct, one-step fine-tuning, since our approach takes advantage of all the domain-specific available unlabeled data. To the best of our knowledge, no previous work has proposed nor evaluated such  two-step fine-tuning approach for AL.\looseness=-1

An important aspect to investigate in this novel two-step fine-tuning pipeline is which textual representation to exploit. In the selection stage of cold-start AL, we can choose among non-fine-tuned embeddings or domain-adapted embeddings (first step of our DoTCAL approach), since no labeled data is available. In the classification stage,  all options (rounded rectangles in Figure 1) are available, but the labeled set is usually small in the beginning of the process, especially with low budgets. It is not completely clear that the same contextual embedding representation that constitutes the state-of-the-art for a fully labeled dataset is the best for AL, especially as input for the selection stage or for the classification stage.\looseness=-1

In this context, we compare contextual embeddings  with other popular text representation paradigms, such as bag-of-words (BoW) and Latent Semantic Indexing (LSI) for the selection as well as the classification phase, considering limited labeling budgets. BoW is still effective in many tasks~\citep{Cunha_2021} while LSI reduces the original term-document matrix to a lower dimensional space, minimizing information loss by exploiting ``more compressed'' latent terms shared by many instances~\citep{LSI}. This approach is suitable in an AL task in which we are interested in selecting only a few representative examples from the whole dataset to label.\looseness=-1

As our \textbf{second contribution}, we evaluate, under different budget scenarios, the effectiveness of these different text representation approaches in each AL stage (selection and classification). More specifically, we aim to answer the following research questions (RQs):\looseness=-1

\noindent \textit{RQ1: How much can we improve AL effectiveness using the proposed two-step fine-tuning (DoTCAL) pipeline?}\looseness=-1

\noindent \textit{RQ2: How does the chosen representation for text impact the selection of highly representative and diverse instances under different budgets?}\looseness=-1

\noindent \textit{RQ3: How does the chosen text representation given as input for the classifier impact classification effectiveness under different budgets?}\looseness=-1

Towards answering these questions, we exploit a state-of-the-art AL strategy in the cold-start scenario named Density Weighted Diversity Based Query Strategy (DWDS) \citep{wang_cswd2021}. This method tackles the selection of informative instances, which is a combinatorial optimization problem, by performing a greed (though extended by a beam-search mechanism) selection of instances with high density (highly representative), discarding those that are similar to other selected instances, also promoting diversity.\looseness=-1

We evaluate this approach using eight ATC datasets under different active labeling budgets. We also vary the representation exploited in each stage of the AL process.  Regarding RQ1, our results show that the 2-step fine-tuning process allows higher classification effectiveness, with gains of up to 33\% in macro-F1 in small budget scenarios and with a lower labeling effort compared to the direct, 1-step fine-tuning. Typically, to reach a given classification effectiveness, our DoTCAL pipeline requires half of the number of labeled instances needed by the traditional 1-step process.\looseness=-1

Answering RQ2, despite some resilience to different representations in the selection stage, the best approach depends on the dataset. Generally, this best approach is the same representation used in the classification stage: BoW/LSI for some datasets and BERT for others.\looseness=-1

Regarding RQ3, although contextual embeddings constitute a state-of-the-art representation for ATC (BERT wins in 5 out of 8 evaluated datasets, for all budget scenarios), BoW and LSI lead to the best classification results in other 3 hard-to-classify datasets. Comparing BoW and LSI, the latter outperforms the former for low-budget scenarios (under 200 labeled instances), confirming our hypothesis that latent dimensions are advantageous when training with small sets of labeled instances since it aggregates information of many instances into fewer dimensions.\looseness=-1

After evaluating BERT contextual embeddings (one of the most popular) general-use, and low-cost (in terms of memory usage) contextual embeddings, we also provide results for a larger robust model, namely RoBERTa \citep{roberta}. Our experiments demonstrate the superiority of the DoTCAL fine-tuning process w.r.t the traditional 1-step fine-tuning, with gains of up to 27\% in Macro-F1, despite some advantages RoBERTa presents for low labeling budgets.

Summarizing, our main contributions are two-fold. First, we propose DoTCAL, a new two-step fine-tuning pipeline for contextual embeddings that outperforms previous one-step SOTA solutions regarding a tradeoff effectiveness-cost (budget size). Second we provide a thorough empirical study  showing the benefits of different text representations in different contexts (datasets) under different budgets in different stages of the AL process.\looseness=-1

\paragraph{Article Organization:} The rest of this paper is organized as follows. Section \ref{sec:related} presents related studies, while Section \ref{sec:statement} formally defines our addressed problem. Section \ref{sec:methodology} describes the evaluated representation and AL approaches, while Section \ref{sec:setup} describes the experimental setup. Section \ref{sec:results} presents our results, while Section \ref{sec:discussion} presents a discussion of the theoretical and practical implications of our study. Finally, Section \ref{sec:conc} provides conclusions and  future work.\looseness=-1
 
%\vspace{-0.4cm}
\section{Related Work} \label{sec:related}

We can divide the various selection criteria that have been exploited in AL strategies into two groups: (1) those that require an initial supervised model, and thus an initial set of labeled instances: and (2) those that do not present such requirement. The former group exploits measures such as model uncertainty and expected model change \citep{Settles2010, emnlp_2020}, while the latter uses measures such as density \citep{Zhu2008, emnlp_2020}, and diversity \citep{Kee2018, Cardoso2017, silva2016compression}. We focus on the second group.\looseness=-1

Those criteria, especially density and diversity, depend on a particular data representation, which is  challenging for high-dimensional data such as text. Among various existing representation approaches, contextual embeddings such as BERT~\citep{devlin2019bert} constitute SOTA approaches for ATC problems.\looseness=-1

Related to our work, ~\cite{emnlp_2020, jacobs2021corr} perform an empirical study of AL techniques for BERT-based classification. However, both of these previous work focus on a non-cold start scenario, in which there is an initially labeled sample (seed). Besides that, while in \cite{jacobs2021corr} no fine-tuning is performed, in \cite{emnlp_2020}, the representations are fine-tuned directly using labeled data, which are usually scarce in AL.  Moreover,  only binary text classification tasks were considered in their evaluation.  We, on the other hand, focus on cold-start AL, while providing a broader evaluation on datasets with two or more classes and propose a fine-tuning pipeline that achieves higher classification effectiveness using less labeled data.\looseness=-1
 
\cite{Zhu2014} evaluated the use of  Latent Semantic Indexing (LSI) in the AL task. The authors propose Global and Local Contribution Ranking (GLCR), which selects  terms (latent dimensions) and documents that make  significant contributions by minimizing information loss. They do not evaluate their strategy under different budgets (maximum amount of selected training instances) and the reported gains in classification effectiveness relative to the original high-dimensional representation are marginal. In our work, we present larger improvements using different ``compression rates'' (i.e., the number of latent dimensions) for different budgets.\looseness=-1

Regarding improvements in contextual embeddings with unlabeled data, the most related work to ours is  \citep{gururangan-etal-2020-dont}, which proposes to exploit unlabeled data from the target domain (domain-adaptive pretraining) and the target task (task-adaptive pretraining) to continue the pretraining process of contextual embeddings, leading to gains in classification effectiveness. They do not consider the active learning scenario nor aspects related to limited budgets,  only fully labeled training data.\looseness=-1

Finally, as we  address relatively small amounts of labeled data to train models, a related research area is few-shot learning (FSL) \citep{fasl, gu-etal-2022-ppt}, which exploits small amounts of labeled data, especially for neural-network solutions. Most current FSL strategies focus on prompt engineering approaches, such as soft prompt tuning \citep{gu-etal-2022-ppt, zhu2023ipm}. Answering which paradigm (prompt or fine-tuning) is the most robust and effective requires further investigation, being out of the scope of this work.  In any case, adapting AL strategies, for example, to select which instances to use for prompt engineering is promising and will be investigated in future work.\looseness=-1

Still related to the prompt-learning paradigm, there is an increasing interest in exploiting Large Language Models (LLMs), due to its high inference power for various NLP tasks \citep{gpt3, zhu2023ipm, suzuki2023ipm}. However, the very-large-scale nature of these models requires a high-cost infra-structure, and one of the goals of AL is to reduce costs. Additionally, the most successful models, such as GPT-3, GPT-4 and ChatGPT, are not open source and are only accessible through APIs provided by companies such as OpenAI\footnote{https://openai.com} and HuggingFace\footnote{https://huggingface.co/}. In this context, privacy issues arise as many application data contain sensitive or confidential information that cannot be submitted through APIs. Thus, the study of smaller-scale models such as BERT and RoBERTa \citep{roberta,tpdr,francca2024representation}, as we perform in this work, is  relevant for the proposal of better cost-effective solutions, including those  with privacy concerns.\looseness=-1

 %\vspace{-0.3cm}
\section{Problem Statement} \label{sec:statement}

We address the pool-based AL problem \citep{Settles2010} for ATC: given a pool $U$ of unlabeled instances (texts), we aim at selecting a set $S \subseteq U$ containing the most informative instances to label. The size of $S$ must not be superior to a given \textit{budget} -- the maximum number of instances one can afford to label. Specifically, We tackle \textit{cold-start} AL, in which there are no previously labeled instances, a common scenario in AL. We divide the AL process into two stages: (i) \textbf{selection stage}, which aims at selecting informative instances, and (ii) \textbf{classification stage}, in which we train a classification model using the instances selected in stage (i), after being actively labeled. We aim at exploring the best text representation approaches for each stage for different scenarios (different labeling budgets). We address single-label, non-hierarchical topic and sentiment classification tasks, covering multi-class classification problems. \looseness=-1

%\vspace{-0.1cm}
 \section{Evaluation Methodology} \label{sec:methodology}

In our evaluation, we compare five text representation approaches: namely, Bag of Words (BoW), Latent Semantic Indexing (LSI), FastText, and BERT-based, in  two versions: no-tuning and fine-tuned. The latter consists of an average pooling over the last four hidden states of the BERT architecture and overall input tokens. \looseness=-1

As classification method, we use Support Vector Machines (SVM) because: (1) it was the best approach in our preliminary experiments, which also included Multilayer Perceptron and KNN-based classifiers, and (2) in previous work, SVM remains a state-of-the-art approach for text classification in various scenarios \citep{Cunha_2021}.\looseness=-1

\subsection{Representation Approaches} \label{sec:representations}

\noindent \textbf{TF-IDF weighted Bag of Words (BoW)}: the most traditional approach, still widely employed \citep{Cunha_2021}.\looseness=-1

\noindent \textbf{Latent Semantic Indexing (LSI)} \citep{LSI}: in this representation, the high-dimensional document-term matrix produced by the BoW-based approach is mapped to a lower-dimensional space using Singular Value Decomposition. Terms that frequently co-occur usually map to the same dimension in the reduced space.\looseness=-1

\noindent \textbf{FastText} \citep{fasttext}: FastText maps words to a dense vector space computed by exploiting word co-occurrences in close positions of the text. It considers ``subword'' information by representing each word as a bag of character n-grams. FastText embeddings are static in the sense that each word produces the same representation, regardless of the semantic context in which it appears. As pre-trained FastText model, we used the 300-dimension English model 
pre-trained using Wikipedia and news datasets.\looseness=-1

\noindent \textbf{BERT and Fine-tuning} \citep{devlin2019bert}: One of the most representative Transformers,  BERT consists of contextual embeddings that, unlike FastText, adapt to the semantic context they are inserted to.  A common practice is fine-tuning BERT models to the target task by exploiting labeled data. In this paper, we propose a  \textit{new fine-tuning pipeline for AL} (Section \ref{sec:fine-tuning}, comparing it with the traditional 1-step approach. It is also possible to exploit contextual embeddings maintaining the original pre-trained weights (no fine-tuning), which may be the only available option  when both unlabeled and labeled data are too scarce or cannot be labeled for whatever reason.\looseness=-1 

In the selection stage of the AL process, we may choose among the untuned embeddings or the embeddings produced in the first step of our new fine-tuning pipeline (recall Figure \ref{fig:al}). 
In the classification stage, we can choose among four options: no-tuning, MLM-only tuning (first step of our approach), ATC-only tuning (i.e., the traditional one-step fine-tuning), and both MLM and ATC tuning (our 2-step fine-tuning).
The evaluation of DoTCAL with different text representations  is also a novel contribution of this work.\looseness=-1

\subsection{DoTCAL: a Two-Step Fine-Tuning Pipeline} \label{sec:fine-tuning}

In the AL context for ATC, only a 1-step traditional fine-tuning pipeline has been exploited. It consists in directly exploiting a (typically small) set of initial labeled data, obtained at the beginning of the AL process, to update model weights using an ATC objective function (e.g., accuracy).\looseness=-1

We propose DoTCAL (Domain and Task Adaptation for Cold-Start Active Learning), a new \textbf{two-step fine-tuning} pipeline. As shown in Figure 1-a: (i) our pipeline enhances the BERT pretraining process using a masked language model (MLM) objective in all available unlabeled data aiming at  adapting the model to the vocabulary of the target domain, and (ii) it further adapts the model using labeled data, that is, the training samples that were actively labeled after being selected by the AL process.\looseness=-1 

More specifically, step (i) updates the language model weights during $e_{\mathit{MLM}}$ epochs, using the Adam optimizer \citep{adam} with an initial learning rate $\lambda_{\mathit{MLM}}$. The MLM objective is defined as the cross-entropy loss on predicting randomly masked tokens \citep{devlin2019bert}. This is done using the unlabeled set $U$, which already provides an initial tuning of the model to the task/domain, since it exploits the own vocabulary of the domain in  the prediction task. In step (ii), the model is further adjusted during $e_{\mathit{ATC}}$ additional epochs, also using the Adam optimizer, with an initial learning rate $\lambda_{\mathit{ATC}}$ and a cross-entropy loss on predicting class labels.\looseness=-1

\subsection{Active Learning Approach} \label{sec:al_approach}

In our study, we explore a state-of-the-art AL strategy named Density Weighted Diversity Based Query Strategy (DWDS) \citep{wang_cswd2021} in which an initial set of labeled data is not available. DWDS is based on two widely used AL selection criteria: (1) density and (2) diversity. The former is measured by the average similarity of an instance $x$ to the top-$k$ most similar instances to $x$ in the unlabeled dataset $U$:
$Density (x, k) = \frac{1}{k} \sum_{i = 1}^k sim(x, x_i)$,  where $\{x_1, x_2, ..., x_k\} \in U$ are the $k$ most similar instances w.r.t. $x$, and $sim$ is a similarity measure, in our case, the cosine similarity.\looseness=-1

\textit{Density} captures the fact  that data instances similar to many other ones lying in denser regions of the space are more representative than those located in low-density regions, which may constitute outliers. \textit{Diversity}, in turn, ensures that a given instance is not very similar to any other instance already selected in the AL process. This is performed to avoid the selection of redundant instances  to stimulate  complementarity. Given a set of selected instances $S$, the diversity of an instance $x$ is defined as the cosine distance (or equivalently, 1 - cosine similarity) between $x$ and the most similar instance in $S$, that is: $Diversity (x, S) = 1 - \max_{s \in S} \{ sim(x, s) \}$. Considering the two measures defined above, we can define our adapted  version of the DWDS algorithm as depicted in Algorithm \ref{alg:dwds}.\looseness=-1

\begin{algorithm}[!h]
\SetKwInOut{Input}{input}
\SetKwInOut{Output}{output}
\small
\caption{DWDS (adapted)}
\label{alg:dwds}
\DontPrintSemicolon

  \Input{Unlabeled data $U$, distance threshold $dist_{min}$, number of instances to be selected \textit{budget}, number of neighbors $k$}
  \Output{Set of selected instances $S$}

   $S \gets \emptyset$ \; $U' \gets U$ \;

 \While{$|S| < $ \textit{budget} \textbf{and} $U' \ne \emptyset$}
 {
    $s \gets argmax_{x \in U'} \{Density(x, k)\}$ \;
     \If{$Diversity(s, S) \ge dist_{min}$}
     {
          $S \gets S \cup \{s\}$ \;
     }
     $U' \gets U' \setminus \{s\}$ \;
 }
\Return $S$ \;
\end{algorithm}

The set of selected instances $S$ is initialized as an empty set (line 1), while the set $U' = U \setminus S$ initially contains all unlabeled instances ($U$). At each iteration of the loop in line 3, the algorithm greedily chooses the instance $s \in U'$ with the highest density measure (line 4). If $s$ is sufficiently complementary to the already selected instances ($S$) according to the diversity measure, given the distance threshold $dist_{min}$, it is included in set $S$ (line 6). The algorithm finishes when $S$ contains the desired number of instances $budget$ or there are no more unlabeled instances that meet the  $dist_{min}$ threshold.\looseness=-1

%In the original work, instead of greedily choosing the highest density instance (line 4 of Algorithm \ref{alg:dwds}), DWDS performs a beam search strategy using a number $b$ of beams ranging from $b$=$1$ up to $b$=$3$. We  exploit only one beam because: (1) the gains exhibited by increasing $b$ are only marginal~\citep{wang_cswd2021} and (2) the costs (execution time) of using b $>$ 1 in our datasets, which are larger in scale than those used  in  \citep{wang_cswd2021}, are prohibitively high.\looseness=-1

 %\vspace{-0.2cm}
\section{Experimental Setup} \label{sec:setup}

\begin{table*}[!h]
\vspace{-0.5cm}
	\small
	\centering	
	\begin{tabular}{c p{4cm} c c c c c c}	
		\toprule
		%\textbf{Name} & \textbf{$|$D$|$} & \textbf{Avg(T)} & \textbf{U(T)} & \textbf{O(T)} & \textbf{$|$C$|$} & \textbf{Max(C)} & \textbf{Min(C)}\\
		
		& \textbf{Name} & \textbf{\#Documents} & \textbf{Avg. \#words} & \textbf{Vocabulary}  & \textbf{\#Classes} & \textbf{Major}    & \textbf{Minor}   \\
		&               &                      & \textbf{per document} & \textbf{size}        &                    & \textbf{class size} & \textbf{class size} \\
		
		\midrule  
		
		\multirow{5}{*}{\rotatebox{90}{Topic}}
		& WebKB (Craven et al'98) & 8199 & 209 & 26398 & 7 & 3705 & 137\\
		& Reuters\footnote{https://martin-thoma.com/nlp-reuters/} & 13327 & 163 & 31668 & 90 & 3964 & 2\\
		& 20NG\footnote{ http://ana.cachopo.org/datasets-for-single-label-text-categorization} & 18846 & 255 & 295605 & 20 & 999 & 628\\
        & ACM \cite{Cunha_2021} & 24897 & 63 & 78149 & 11 & 6562 & 63\\
        %& Books & 33594 & 268 & 242297 & 8 & 4934 & 1226\\
        %& DBLP & 38128 & 142 & 136296 & 10 & 9746 & 1414\\
        & AGNews \cite{agnews}& 127600 & 40 & 171526 & 4 & 31900 & 31900\\
        %& Sogou & 510000 & 542 & 510631 & 5 & 102000 & 102000\\
        \midrule
        \multirow{3}{*}{\rotatebox{90}{Senti.}} 
        
        %& Aisopos & 278 & 16 & 1822 & 2 & 159 & 119\\
        %& VaderNYT & 4946 & 18 & 15671 & 2 & 2742 & 2204\\
        & Yelp2L\cite{mendes2020keep} & 5000 & 129 & 37924 & 2 & 2500 & 2500\\
        %& SST2 & 9613 & 17 & 17515 & 2 & 4963 & 4650\\
        & VaderMovie & 10568 & 19 & 21952 & 2 & 5326 & 5242\\
        %& PangMovie & 10662 & 19 & 21383 & 2 & 5331 & 5331\\
        & IMDB & 348415 & 297 & 112997 & 10 & 63233 & 12836
        %& Yelp2015 & 700000 & 137 & 794635 & 5 & 140000 & 140000\\
        \\
		%Webkb & 8199 & 208.8 & 26398 & 32.9 & 7  & 3705 & 137  \\
		%20ng & 18846 & 296.4 & 349614 & 8.2 & 20  & 999 & 628  \\
		%Acm & 24897 & 63.5 & 100702 & 4.0 & 11  & 6562 & 63  \\
		%reut &  &  &  &  &   &  &   \\ \hline
        %Debate & 1979 & 14.3 & 6894 & 4.0 & 2  & 1249 & 730  \\
        %Ss\_rw & 705 & 66.9 & 9543 & 3.3 & 2  & 484 & 221  \\
        %Yelp R & 5000 & 131.7 & 40056 & 11.7 & 2  & 2500 & 2500  \\
        %Sst-2 & 68221 & 9.5 & 15756 & 30.3 & 2  & 38013 & 30208  \\
        %Aisopos & 278 & 15.2 & 2028 & 2.1 & 2  & 159 & 119  \\
        %Nikolaos  & 727 & 16.1 & 2252 & 4.5 & 2  & 409 & 318  \\

%Pang movie & 10662 & 21.0 & 21418 & 10.0 & 2  & 5331 & 5331  \\

		\bottomrule 
	\end{tabular}
 \caption{Dataset Statistics.}
\label{tbl:datasets_stats}
\end{table*}

We consider a total of \textbf{eight} ATC datasets divided into two groups~\citep{cunha2023comparative}: (i) \textbf {topic classification} and (ii) \textbf {sentiment analysis}. For topic classification, we exploited \textbf {five} benchmark datasets: 20 Newsgroups (20NG), ACM DL, AGNews, Reuters and, WebKB. For sentiment analysis, we consider other \textbf{three} benchmark datasets: IMDB Reviews, Vader Movie Reviews - VaderMovie, and YelpReviews. All these datasets have been used as benchmarks by most works in ATC and are detailed in Table~\ref{tbl:datasets_stats}. \looseness=-1
We can observe ample diversity in many aspects of these datasets regarding size, dimensionality (i.e., vocabulary size), number of classes, and density (the average number of words per document). In addition, the class distributions of these datasets present different levels of skewness, ranging from highly balanced (AGNews) to very skewed.\looseness=-1

The experiments in the smaller datasets (containing 100k documents or less) were executed using a 10-fold cross-validation procedure. For the larger datasets (AGNews and IMDB), we used 5-fold cross-validation due to the cost of the procedure~\citep{cunha2023sigir}. For each fold, the AL pool ($U$) corresponds to the training set. Only the labels of the selected instances ($S$) are considered to train classifiers in the classification stage, simulating a real active labeling process. For other details on parameter tuning, please check Appendix A. To compare the average results of our cross-validation experiments -- considering Macro-F1~\citep{Cunha_2020} as evaluation metric due to dataset imbalance --
we assess statistical significance by employing a paired Wilcoxon test with 95\% confidence. \looseness=-1

All models have been evaluated using the same hardware configuration: an AWS p3.2xlarge instance with eight vCPUs, 64 GB of memory, 1x NVIDIA V100 GPU.\looseness=-1

\begin{figure*}[!ht]

     \centering
     \begin{subfigure}[t]{0.32\textwidth}
         \centering
         \includegraphics[width=\textwidth]{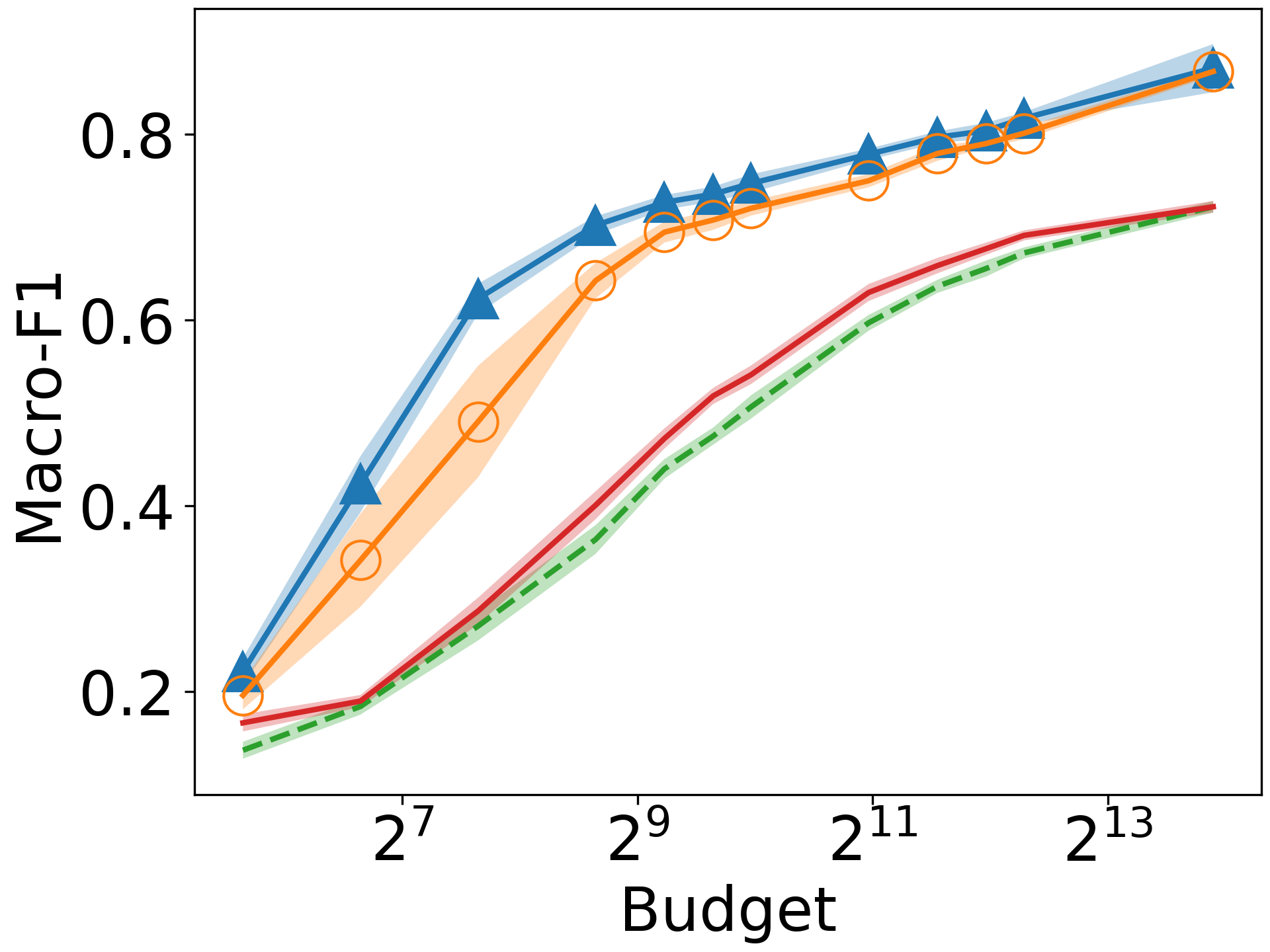}
         \caption{20NG}
         \label{a}
     \end{subfigure}
\hfill
     \begin{subfigure}[t]{0.32\textwidth}
         \centering
         \includegraphics[width=\textwidth]{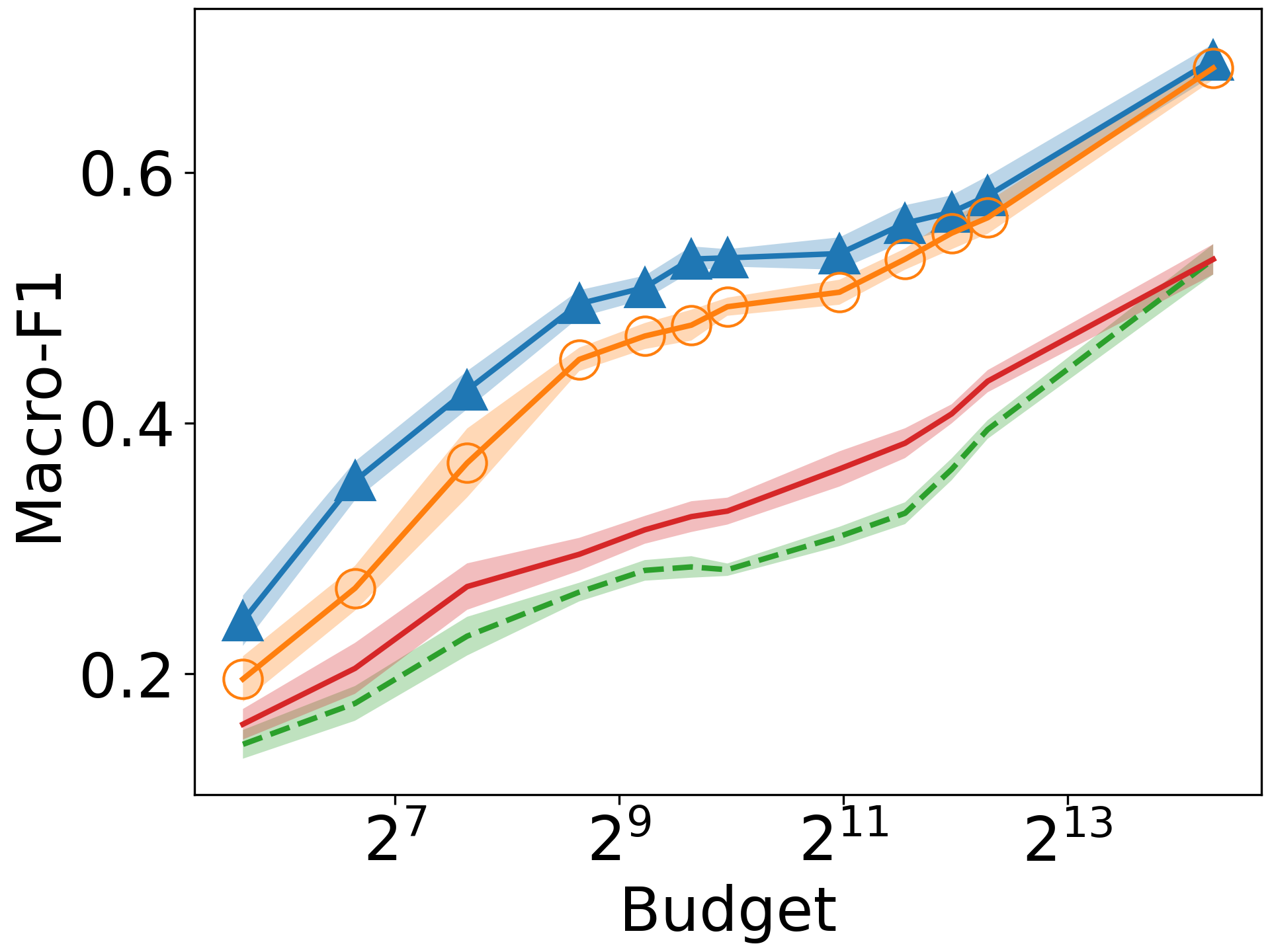}
         \caption{ACM}
         \label{b}
     \end{subfigure}
\hfill
     \begin{subfigure}[t]{0.32\textwidth}
         \centering
         \includegraphics[width=\textwidth]{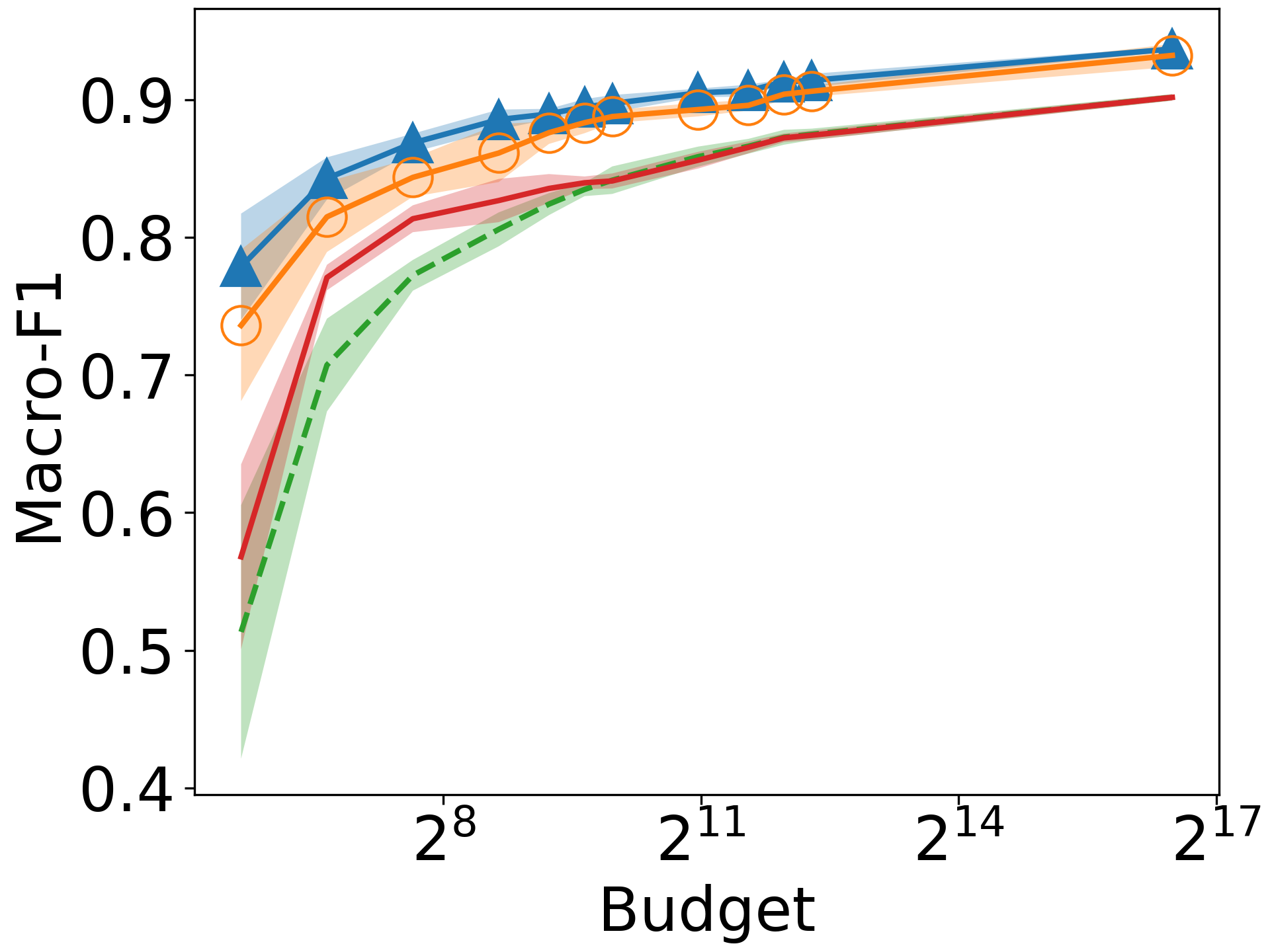}
         \caption{AGNews}
         \label{c}
     \end{subfigure}
\hfill
     \begin{subfigure}[t]{0.32\textwidth}
         \centering
         \includegraphics[width=\textwidth]{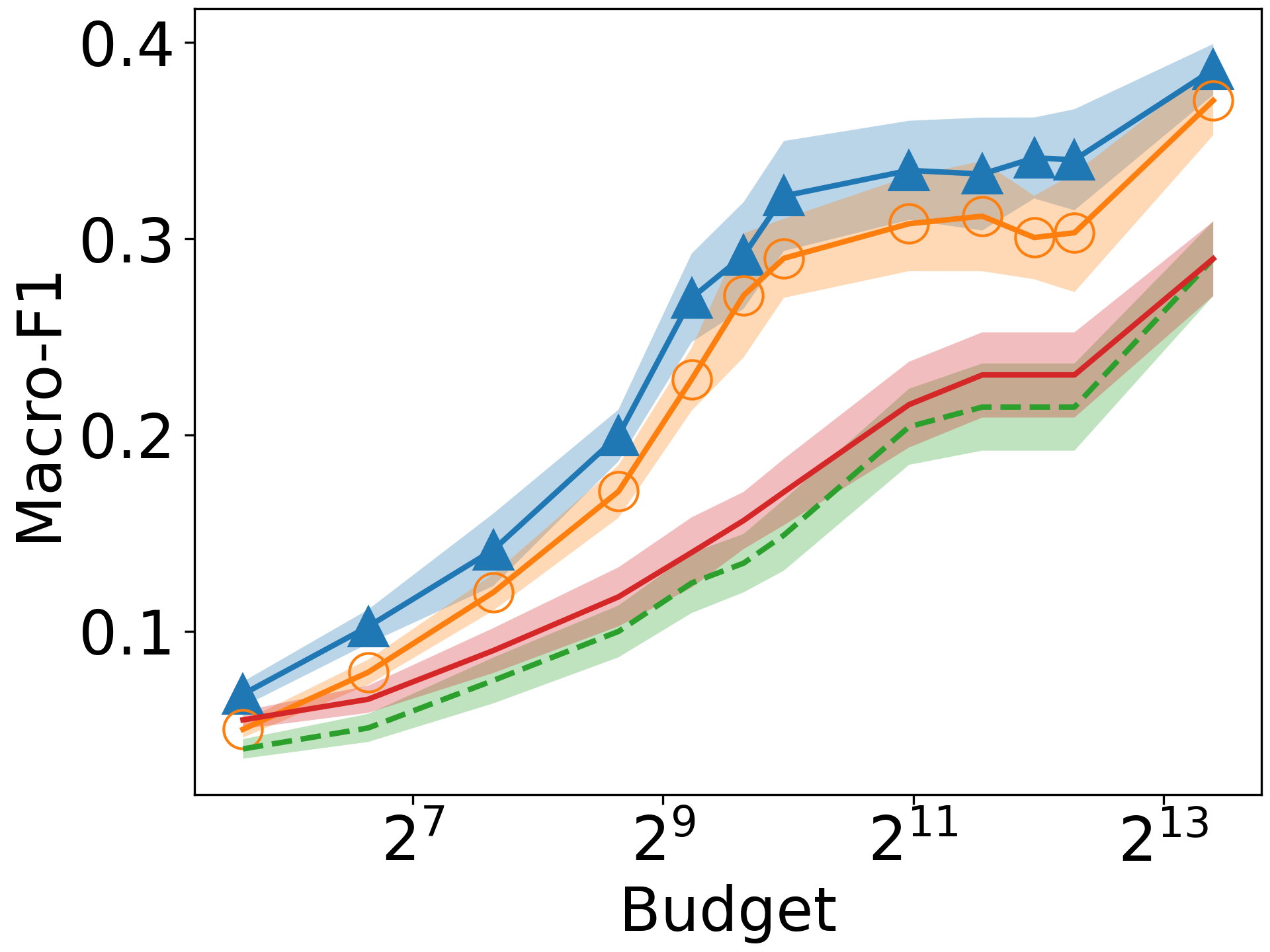}
         \caption{Reuters}
         \label{d}
     \end{subfigure}
\hfill
     \begin{subfigure}[t]{0.32\textwidth}
         \centering
         \includegraphics[width=\textwidth]{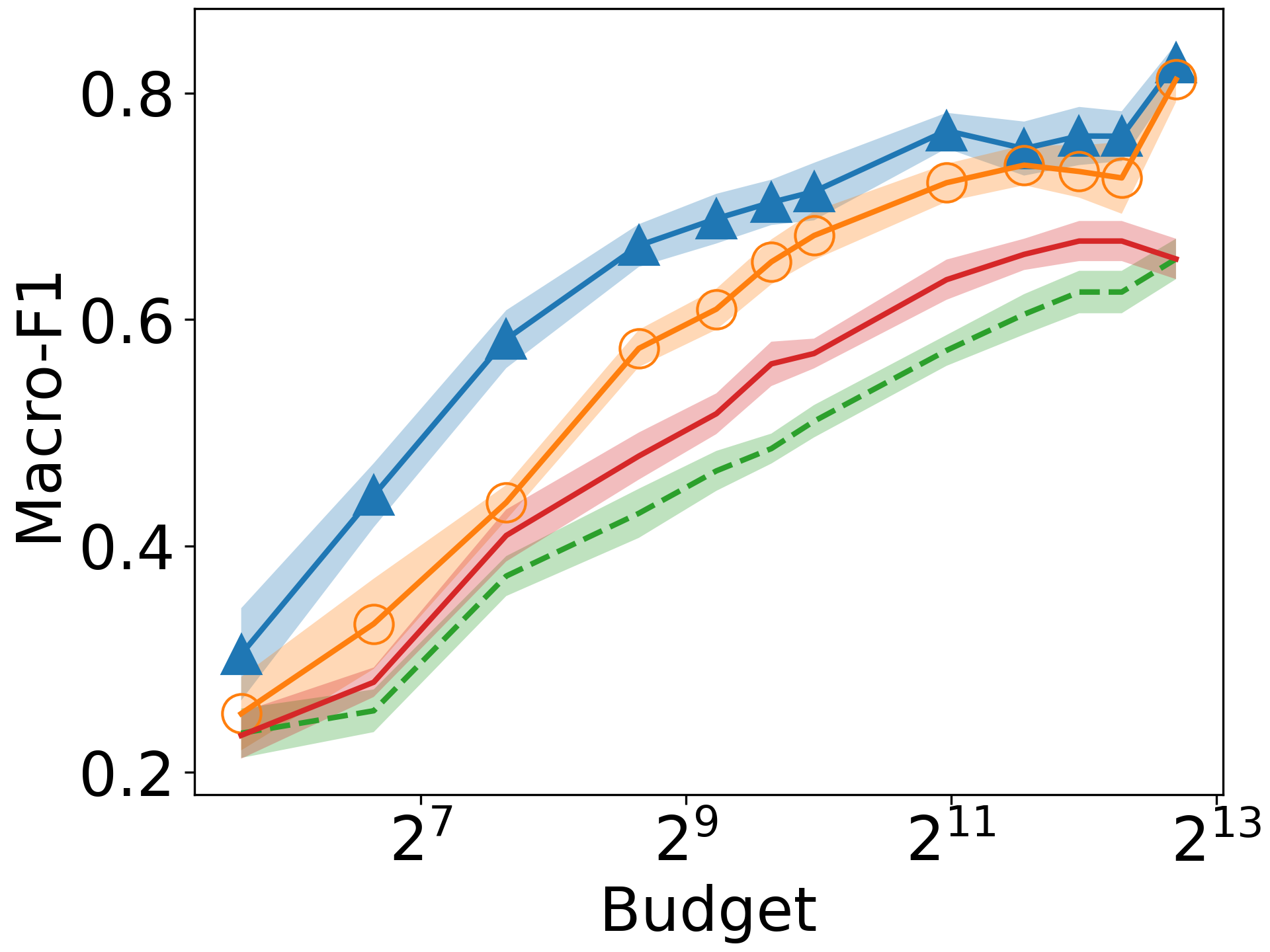}
         \caption{WebKB}
         \label{e}
     \end{subfigure}
\hfill
     \begin{subfigure}[t]{0.32\textwidth}
         \centering
         \includegraphics[width=\textwidth]{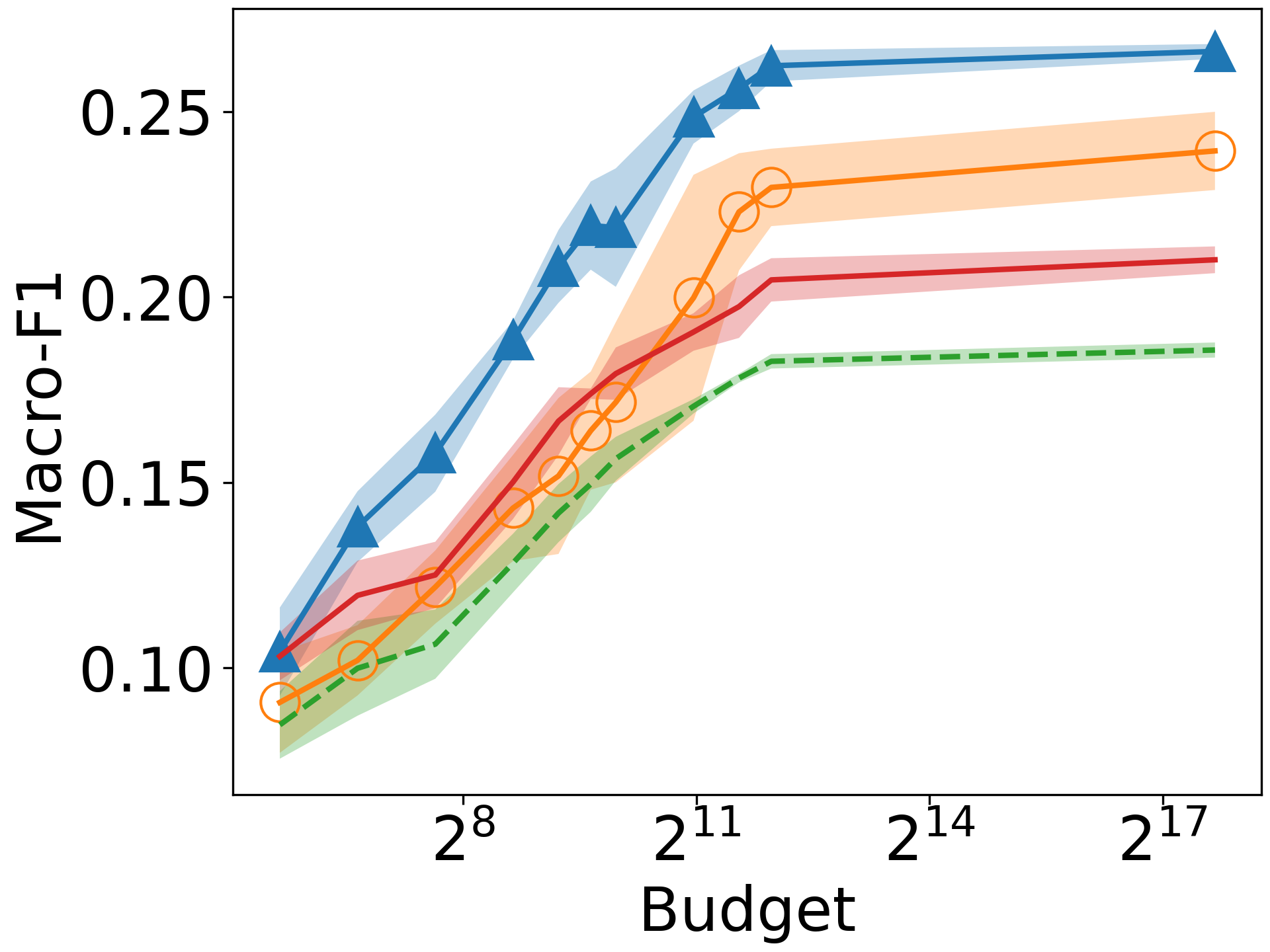}
         \caption{IMDB Reviews}
         \label{f}
     \end{subfigure}
\hfill
     \begin{subfigure}[t]{0.32\textwidth}
         \centering
         \includegraphics[width=\textwidth]{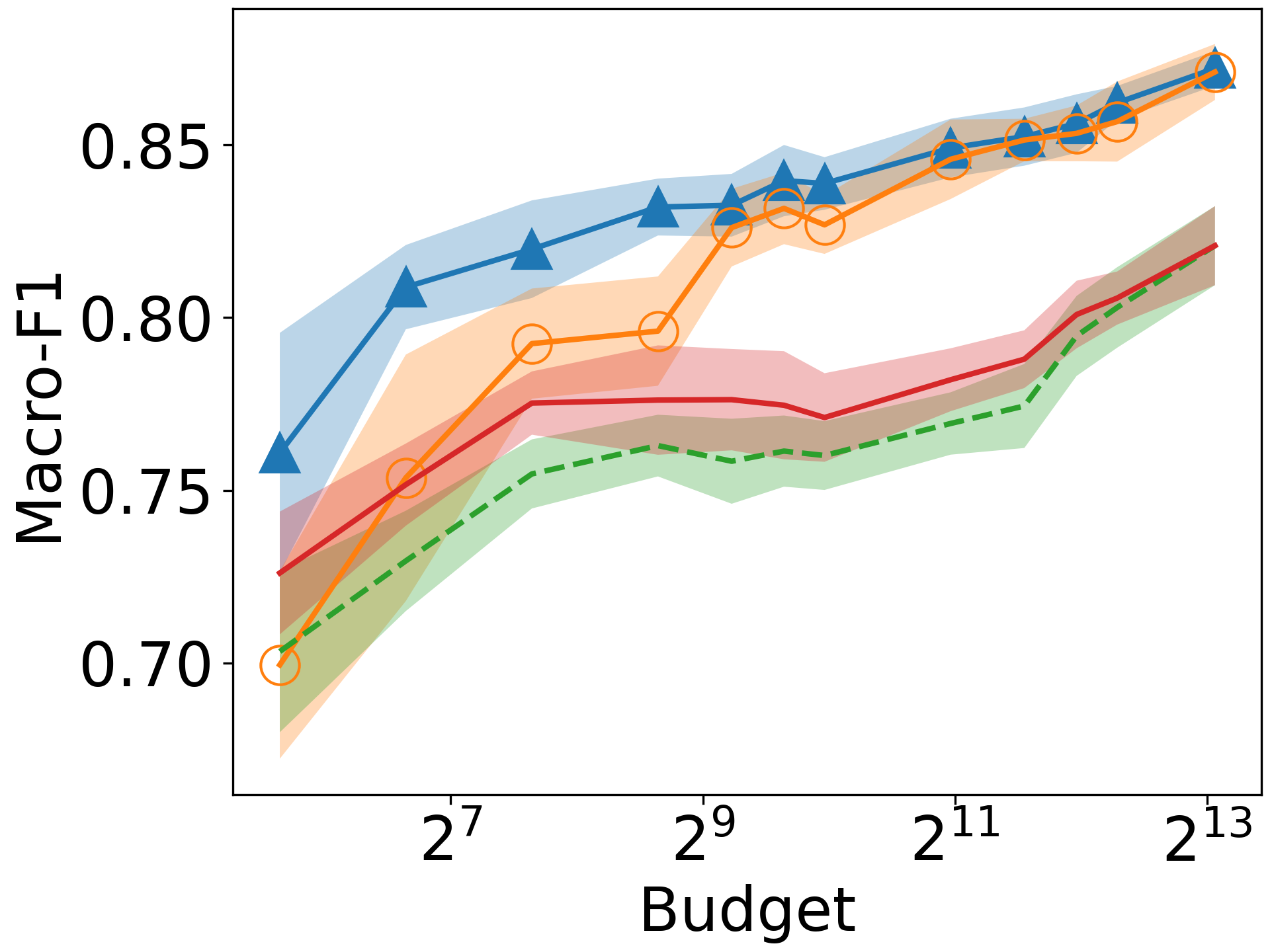}
         \caption{VaderMovie}
         \label{g}
     \end{subfigure}
\hfill
     \begin{subfigure}[t]{0.32\textwidth}
         \centering
         \includegraphics[width=\textwidth]{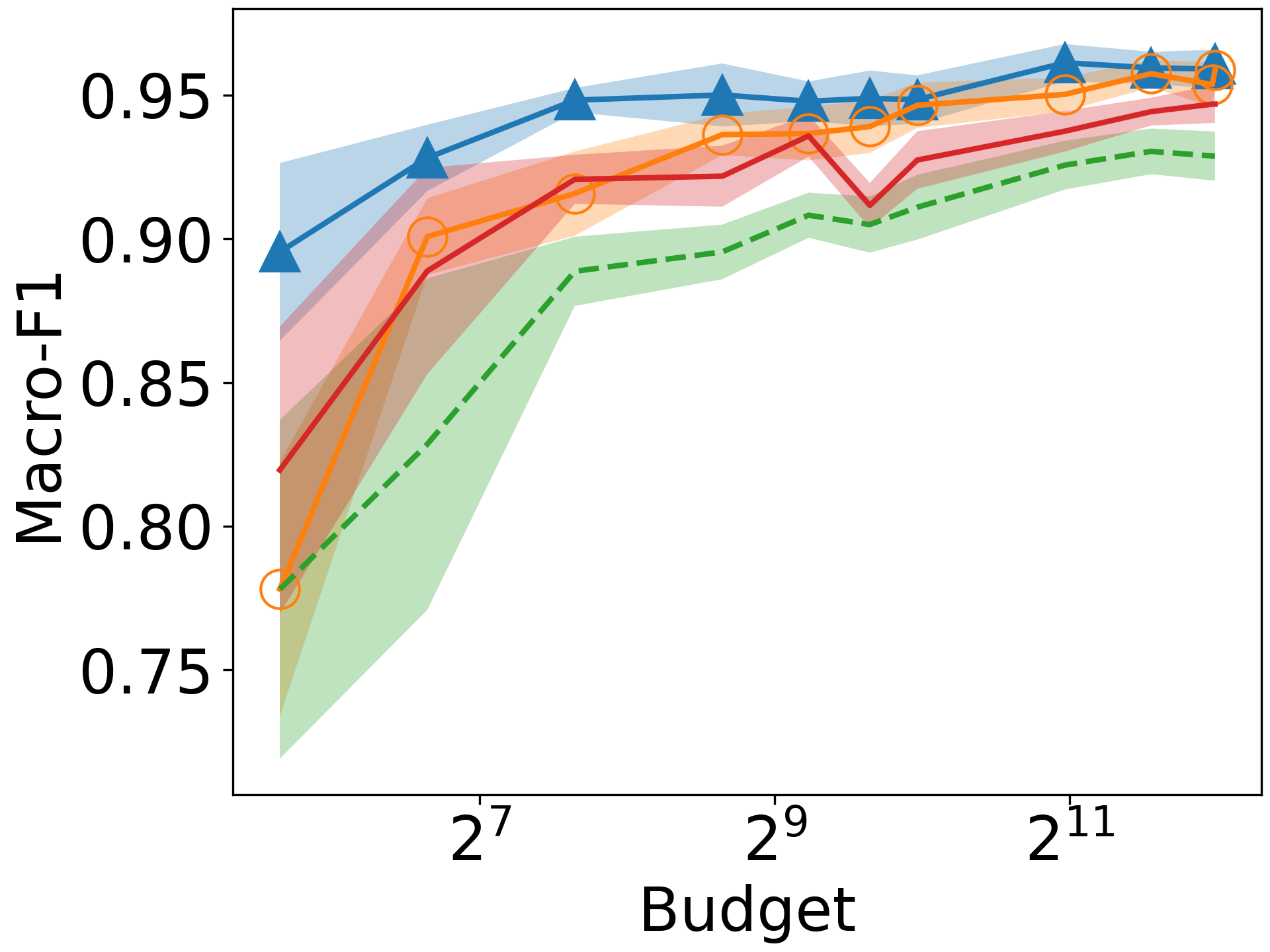}
         \caption{YelpReviews}
         \label{h}
     \end{subfigure}
\hfill
     \begin{subfigure}[t]{0.32\textwidth}
         \centering
         \includegraphics[width=\textwidth]{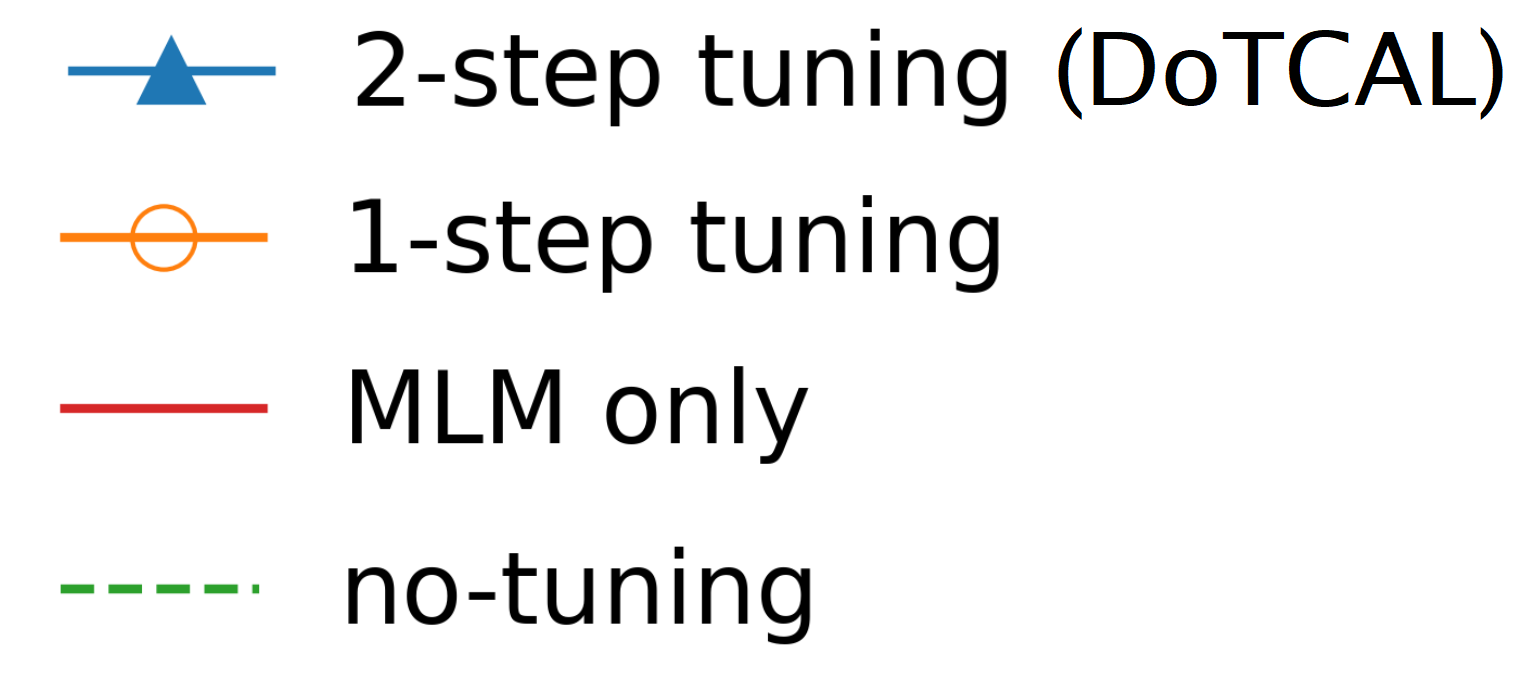}
         %\caption{YelpReviews}
         %\label{a}
     \end{subfigure}
\hfill
     \caption{\footnotesize{Macro-F1 for the BERT representation with different fine-tuning: DoTCAL, the traditional 1-step, MLM only (i.e., applying only the first step of our approach) and no fine-tuning. 95\% confidence intervals shown in shaded areas.}}
    \label{fig:res}
\end{figure*}

%\newpage
%~
%\newpage

%\vspace{-0.3cm}
\section{Experimental Results} \label{sec:results}

In this section, we answer the research questions (RQs) stated in Section \ref{sec:intro}. 
We first discuss the results of different contextual embeddings fine-tuning pipelines for AL, in comparison with our proposed DoTCAL approach (Sec.~\ref{sec:res_rq1}). Then, we analyze the impact of different text representation approaches on AL selection (Sec.~\ref{sec:res_rq2}) and classification (Sec.~\ref{sec:res_rq3}) stages. Finally, we analyze the impact and benefits of LSI in AL (Sec.~\ref{sec:lsi}), and generalize the results of our strategies to another language model, namely RoBERTa (Sec.~\ref{sec:roberta}). \looseness=-1

\vspace{-0.4cm}
\subsection{RQ1: DoTCAL Effectiveness} \label{sec:res_rq1}

We evaluate DoTCAL for BERT-based representations in the AL context, comparing it with the traditional one-step fine-tuning, as well as with other possible paths in the fine-tuning pipeline: the absence of fine-tuning and the application of only the first stage of our approach (MLM only). Figure \ref{fig:res} shows, considering different amounts of labeled data (budget), macro-F1 results for the AL classification stage, for each of these possible paths. To obtain these results, we fixed the representation in the selection stage as BoW (we evaluate other representations' results in the selection stage in Section~\ref{sec:res_rq2}).\looseness=-1

Our first observation is that, even with relatively small amounts of labeled training data (small budgets), any fine-tuning approach greatly outperforms the absence of fine-tuning. Thus, similarly to the fully labeled classification scenario~\citep{gururangan-etal-2020-dont}, a fine-tuning approach is essential in the AL process.

Now we compare the different fine-tuning approaches. The first step of our DoTCAL approach -- MLM only tuning --  that does not require any labeled data, outperforms untuned embeddings with gains of up to 37\% in macro-F1, for any budget. However, the most effective fine-tuning approaches are those which exploit (at least some) labeled data: DoTCAL and the 1-step traditional approach. DoTCAL  produces significant gains in macro-F1 over the 1-step approach for all datasets, particularly for relatively small budgets. Considering \textit{budget}=200, our approach outperforms the  1-step process with a maximum gain of 33\% in WebKB dataset and average gains of 17\% over all datasets.\looseness=-1

From another perspective, we note that DoTCAL requires a significantly lower amount of labeled data to achieve the same effectiveness level of the 1-step process. For instance, for all datasets, the 1-step fine-tuning approach requires at least 800 labeled instances to reach the same macro-F1 DoTCAL provides with only 400 instances. We observe similar results for other amounts of labeled instances: DoTCAL requires approximately half of the labeling effort required by the 1-step fine-tuning, considering budgets under 1000. \looseness=-1 % instances.\looseness=-1

For some datasets, as the budget approaches the whole training dataset size (right-most points in the graphic), the differences tend to reduce, as the 1-step fine-tuning is  effective when using a large amount of labeled data. However, even in the ``entire labeled dataset'' scenario, we observe non-negligible gains, in macro-F1, of our DoTCAL fine-tuning pipeline over the traditional approach, for two datasets: Reuters (4.3\%) and IMDB Reviews (11.2\%). In these datasets, a more robust fine-tuning pipeline is essential to allow better discrimination among a large number of classes (90 topics in Reuters and 10 sentiment levels in IMDB Reviews). Thus, even in scenarios with high availability of labeled data, continuing the BERT pre-training oriented with a masked language model objective in the target domain can improve classification effectiveness ~\citep{gururangan-etal-2020-dont}. Finally, in some datasets our DoTCAL pipeline reaches the same effectiveness of a fully labeled dataset tuned with the 1-step with a much reduced effort, for 
 instance, only 2\% of the IMDB labeled data and only 49\% of YelpReviews.\looseness=-1

Answering RQ1, DoTCAL allows higher classification effectiveness (up to 33\%)  with a significantly lower labeling effort (half) compared to the traditional 1-step fine-tuning. \looseness=-1

\subsection{RQ2: Impact on the Selection Stage} \label{sec:res_rq2}

We compare the effectiveness of different representation approaches in the selection stage of AL, aiming to answer RQ2. The compared representation approaches are: (1) the best BERT-based representation (using DoTCAL fine-tuning pipeline) (2) Bag-of-Words (BoW), (3) Latent Semantic Indexing (LSI), and (4) FastText. \looseness=-1

Table~\ref{tab:res} shows average macro-F1 results varying (1) the representation employed to select instances (1st column) and (2) the representation employed to classify (2nd column). Each block corresponds to a representation approach employed in the selection stage. The best results (and statistical ties) per block are shown as shaded entries, while the best overall results are shown in \textbf{bold}. These results are obtained setting a \textit{budget} of 200 instances. Similar findings can be obtained for other budgets, results we omit due to space limitations.\looseness=-1

We first analyze macro-F1 results fixing  the classification stage representation as BoW (first line in each block of Table \ref{tab:res}). Considering this setting, for all datasets, BERT is at most tied with the best representation approach (either BoW or LSI), which offers gains of up to 207\% in macro-F1 (e.g., WebKB dataset) over BERT. One may argue that this occurs because BoW and LSI representations are more similar to the representation we are fixing as input to the classification stage (BoW). However, even fixing BERT as input to the classification stage (fourth row) and varying the representation for the selection stage, BoW and LSI are still the best approaches in 3 out of 8 datasets (20NG, Reuters and WebKB, with gains of up to 28\% over BERT) in the selection stage. This indicates that, although BERT constitutes a state-of-the-art approach as input for the classification stage, it is not always superior as input for selecting representative instances in AL, possibly due to limitations of BERT to learn with few data.\looseness=-1

\begin{table*}[!ht]
    	\centering	
 \small
 %\tiny
	\begin{tabular}{l l c c c c c c c c}	
		\toprule
	
		\multicolumn{2}{c}{Representation} & \multicolumn{8}{c}{} \\
	    Selection Stage & Classific. Stage & 20NG  &   ACM  &   Reuters  &  AGNews &   WebKB & IMDB  &   VaderMov.  &   YelpRev. \\ 
	    
	    \midrule

          BoW & BoW & 0.595 & 0.397 & \cellcolor{blue!25}0.162 & 0.758 & 0.399 & 0.083 & 0.605 & 0.858 \\ 
          BoW & LSI & \cellcolor{blue!25}\textbf{0.621} & \cellcolor{blue!25}\textbf{0.413} & \cellcolor{blue!25}0.169 & 0.769 & 0.417 & 0.092 & 0.608 & 0.863 \\ 
     BoW & FastText & 0.160 & 0.159 & 0.028 & 0.790 & 0.173 & 0.050 & 0.649 & 0.725 \\ 
         BoW & BERT & 0.406 & 0.313 & 0.106 & \cellcolor{blue!25}0.815 & \cellcolor{blue!25}0.475 & \cellcolor{blue!25}0.142 & \cellcolor{blue!25}\textbf{0.775} & \cellcolor{blue!25}\textbf{0.930} \\
		 
\hline
		 
          LSI & BoW & 0.537 & 0.347 & 0.179 & 0.766 & 0.451 & 0.080 & 0.599 & 0.857 \\ 
          LSI & LSI & \cellcolor{blue!25}0.569 & \cellcolor{blue!25}0.356 & \cellcolor{blue!25}\textbf{0.196} & 0.785 & 0.476 & 0.088 & 0.594 & 0.861 \\ 
     LSI & FastText & 0.196 & 0.155 & 0.028 & 0.786 & 0.330 & 0.046 & 0.680 & 0.750 \\ 
         LSI & BERT & 0.459 & 0.318 & 0.123 & \cellcolor{blue!25}\textbf{0.838} & \cellcolor{blue!25}\textbf{0.540} & \cellcolor{blue!25}0.165 & \cellcolor{blue!25}\textbf{0.764} & \cellcolor{blue!25}\textbf{0.930} \\ 
		 
\hline
		 
     FastText & BoW & 0.401 & 0.301 & 0.152 & 0.742 & 0.420 & 0.113 & 0.604 & 0.816 \\ 
     FastText & LSI & \cellcolor{blue!25}0.442 & \cellcolor{blue!25}0.322 & \cellcolor{blue!25}0.161 & 0.766 & 0.453 & 0.127 & 0.203 & 0.280 \\ 
FastText & FastText & 0.121 & 0.153 & 0.023 & 0.784 & 0.270 & 0.063 & 0.647 & 0.810 \\ 
    FastText & BERT & 0.350 & 0.279 & 0.106 & \cellcolor{blue!25}\textbf{0.847} & \cellcolor{blue!25}0.477 & \cellcolor{blue!25}\textbf{0.174} & \cellcolor{blue!25}0.747 & \cellcolor{blue!25}\textbf{0.921} \\

\hline

         BERT & BoW & 0.439 & 0.342 & 0.153 & 0.755 & 0.147 & 0.089 & 0.603 & 0.792 \\ 
         BERT & LSI & \cellcolor{blue!25}0.483 & \cellcolor{blue!25}0.364 & \cellcolor{blue!25}0.165 & 0.781 & 0.178 & 0.099 & 0.599 & 0.815 \\ 
    BERT & FastText & 0.151 & 0.163 & 0.019 & 0.797 & 0.089 & 0.047 & 0.672 & 0.433 \\ 
        BERT & BERT & 0.400 & 0.326 & 0.110 & \cellcolor{blue!25}\textbf{0.848} & \cellcolor{blue!25}0.421 & \cellcolor{blue!25}0.158 & \cellcolor{blue!25}\textbf{0.758} & \cellcolor{blue!25}\textbf{0.934} \\

		\bottomrule
	\end{tabular}
 \caption{\footnotesize{Average Macro-F1 for each pair (representation  to select instances, representation to classify). Each block corresponds to a representation approach employed in the selection stage of the AL process. Best results (and statistical ties) per block are shown as shaded entries. Best overall results in bold. Budget is set to 200 instances. Similar conclusions hold for other budgets.}}
 \label{tab:res}
\end{table*}

Now we analyze results when fixing the best representation used as input for the classifier (shaded entries in Table \ref{tab:res}), and varying the representation employed in the selection stage. Focusing first on AGNews, WebKB, IMDBReviews, VaderMovie and YelpReviews datasets, for which the best classifier is BERT-based, various representation approaches employed in the active selection stage achieve similar results. For the AGNews dataset, for example, LSI, FastText, and BERT are statistically tied for the selection stage, while BoW macro-F1 results are only 4\% lower than results for the best approach. For YelpReviews, all approaches are statistically tied when using BERT as classifier, probably due to the high classification effectiveness in this dataset. For the other 3 datasets (20NG, ACM, Reuters), BoW and LSI are the best alternatives in the AL selection stage.\looseness=-1

Thus, answering RQ2, despite some resilience to different representation approaches in the selection stage, the best approach to be used in the selection depends on the dataset. Generally, this best selection approach  is  the same representation approach used in the classification stage: BoW and LSI for some datasets and BERT for others.\looseness=-1

\subsection{RQ3: Impact on Classification Stage} \label{sec:res_rq3}

In this section, we compare the effectiveness of different representation approaches in the classification stage of AL, aiming to answer RQ3 in an enviroment of low budget. For this, we use again  Table \ref{tab:res}. We note that, regardless of the representation we fix in the selection stage, BERT is the best representation in the classification stage for most (5 out of 8) datasets: AGNews, WebKB, IMDB Reviews, VaderMovie, and YelpReviews. For those datasets, BERT provides gains ranging from 8\% to 54\% over the runner-up approach. The second best representation, on the classification stage, varies among BoW, LSI and FastText depending on the dataset.\looseness=-1

In other  3 datasets -- 20NG, ACM and Reuters --  BoW and LSI are the most effective. In particular, in these datasets, LSI produced gains over  BERT ranging from 12\% up to 59\%. Notice that these three datasets are the ones with the largest number of classes. This higher classification difficulty, combined with the low number of labeled instances (low budget), may be harming the BERT classification capabilities. Interestingly, although LSI compresses the sparse BoW space into a lower dimensional space, at the cost of some potential information loss, LSI outperforms BoW in this low-budget scenario, a result that is inspected further in Appendix D. That analysis shows that LSI compression seems to help emphasizing what is important (or not) for the sake of selecting representative and diverse instances.\looseness=-1

Thus, answering RQ3, although contextual embeddings constitute a state-of-the-art representation for classification (and in fact, BERT wins in 5 out of 8 of our evaluated datasets, for all budget scenarios), traditional approaches such as BoW and LSI may lead to superior  results  in the classification stage of AL, especially for low-budget -- highly-difficult scenarios. In particular, in low-budget AL scenarios, unlike scenarios with many labeled data, it is interesting to exploit LSI using only a few (e.g., 50-100) latent dimensions, which produce gains in comparison with full-dimensional BoW representations.\looseness=-1

\subsection{Experiments with the RoBERTa language model} \label{sec:roberta}

\begin{figure*}[ht]
\hfill
     \begin{subfigure}[t]{0.32\textwidth}
         \centering
         \includegraphics[width=\textwidth]{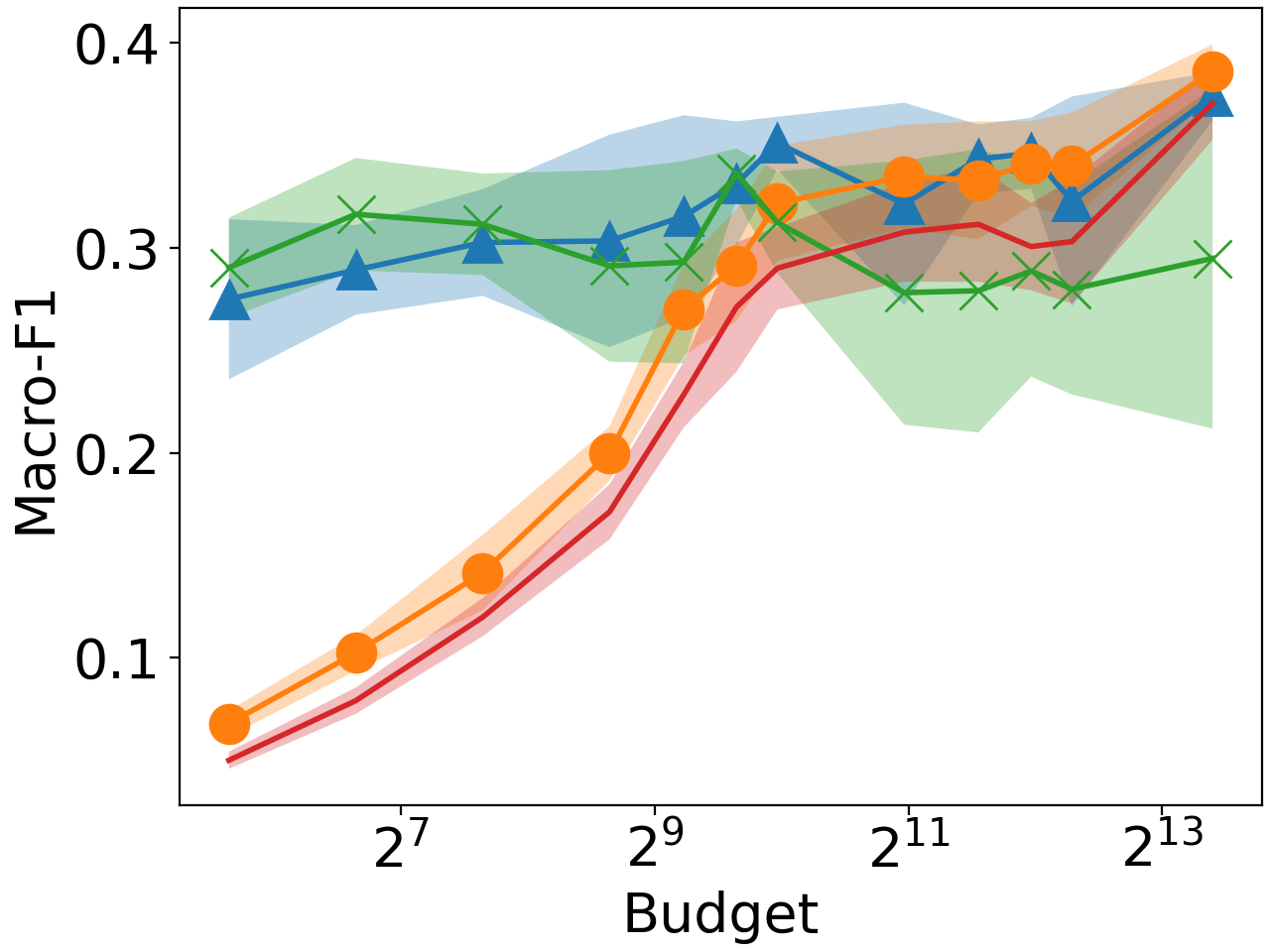}
         \caption{Reuters}
         \label{a}
     \end{subfigure}
\hfill
     \begin{subfigure}[t]{0.32\textwidth}
         \centering
         \includegraphics[width=\textwidth]{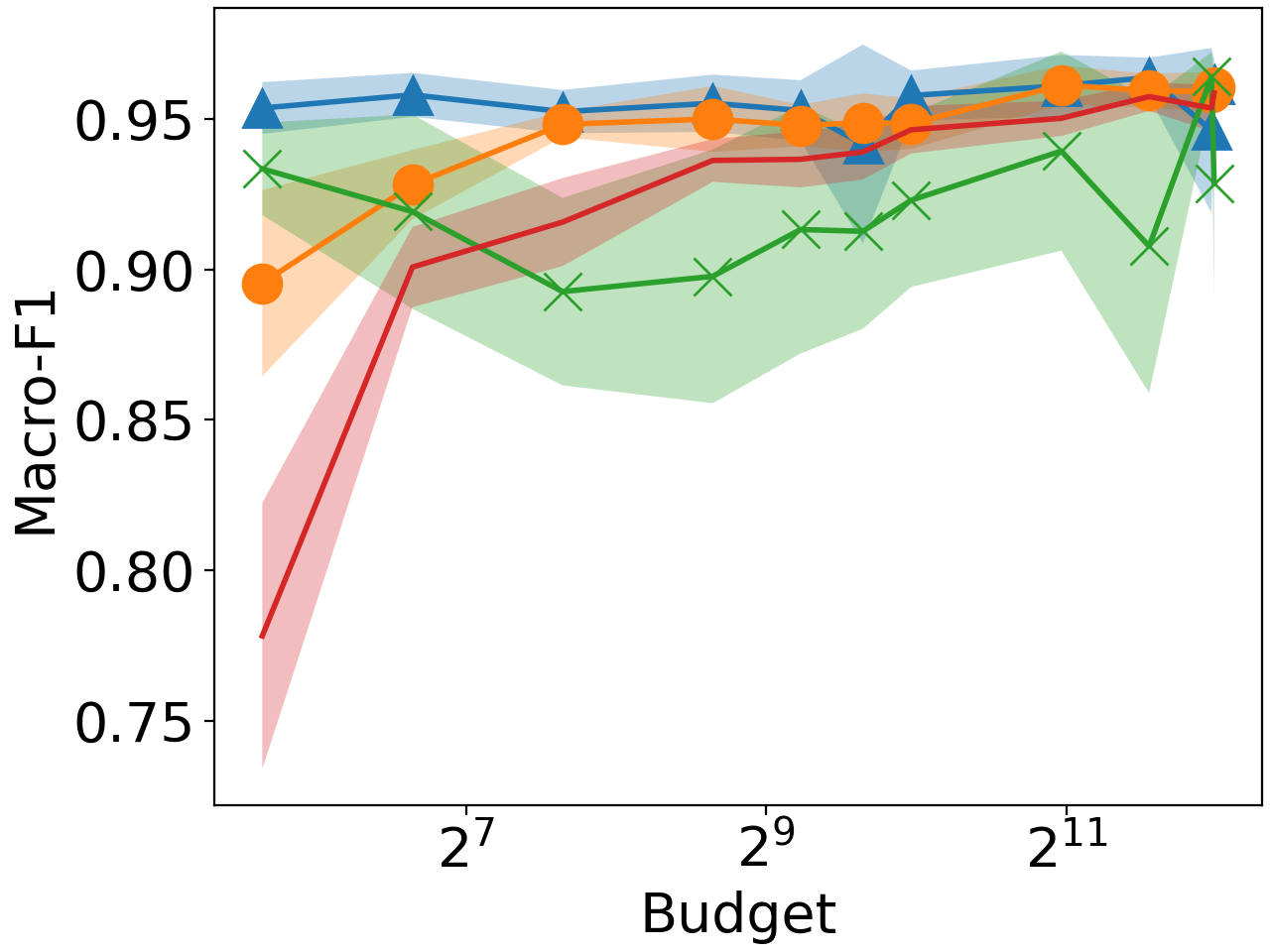}
         \caption{YelpReviews}
         \label{b}
     \end{subfigure}
\hfill
     \begin{subfigure}[t]{0.32\textwidth}
         \centering
         \includegraphics[width=\textwidth]{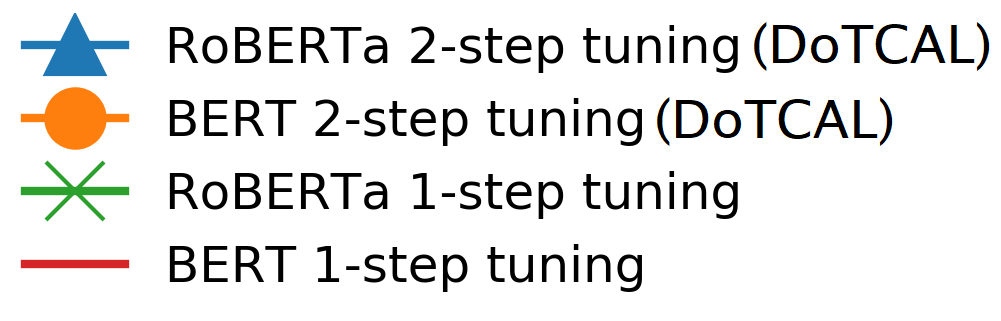}
         %\caption{}
         %\label{}
     \end{subfigure}
     
     \caption{\footnotesize{Macro-F1 for BERT and RoBERTa representations for different fine-tuning approaches and budget sizes. 95\% confidence intervals shown in shaded areas.}}
    \label{fig:res_roberta}
\end{figure*}

In this section, we compare BERT results with another state-of-the-art text representation, namely RoBERTa, showing that our conclusions generalize to other small to medium scale language models. For the \textit{topic} classification datasets evaluated by \cite{claudio_ipm}, including those we evaluate in this article, BERT and RoBERTa results present no statistically significant difference. However, unlike the topic datasets, for various sentiment classification datasets (5 out of the 8 datasets studied in \citep{claudio_ipm}), RoBERTa outperforms BERT with a small margin (under 4\% in Macro-F1). %ESSA FRASE ESTA INCONSISTENTE. SE NAO HA DIFERENCA ESTATISITCA, UM NAO PODE SUPERAR O OUTRO. O empate estatistico é apenas para os datasets de tópicos. Reescrevi para deixar isso mais claro.}
 Here we focus our evaluation on two datasets: (1) the topic classification dataset Reuters; and (2)  the sentiment classification dataset Yelp Reviews,  for which BERT and RoBERTa results tend to be more  distinguishable, in our label scarceness scenario (differences of up to 305\% in Macro-F1).

Figure \ref{fig:res_roberta} shows Macro-F1 results for BERT and RoBERTa based representations for the two fine-tuning approaches (1-step and DoTCAL) and different amounts of labeled data (budget). 

Our first observation is that, unlike  the full-labeled scenario (right-most points in each graphic), in which BERT and RoBERTa results are statistically tied, in low-budget AL scenarios, RoBERTa greatly outperforms BERT with gains of up to 305\% in Macro-F1. This probably due to the fact that RoBERTa was trained in larger collections of data and its language model is at least twice the size of BERT model allows it to be more robust in label-scarce scenarios.

Our second observation is that, even using a more robust language model as RoBERTa, DoTCAL still leads to gains of up to 27\% in Macro-F1, a result that is more evident for all budgets in Yelp Reviews dataset. Even when statistically tied (for all budgets of the Reuters dataset), DoTCAL has the advantage of being more stable, that is, results tend to present a smaller variability: the deviations from the mean are 43\% up to 63\% larger for the 1-step fine tuning than for DoTCAL, on average.

Finally, we also note that increasing the amount of labeled data for the sentiment classification task, specially when using RoBERTa-based representations, little contributes to increase classification effectiveness, %in our cold-start AL scenario,
 which can be explained by two reasons: (1) in this dataset there are only two classes, naturally requiring fewer training examples, and (2) the aforementioned much larger size of RoBERTa when compared to BERT.

%\textbf{E SOBRE O OUTRO CENARIO? FAZ DIFERENCA? TAH MAMNEIRA QUE ESTAH ESCRITO ESTA PUXANDO PRA BAIXO. Não avaliamos cenários não-cold start, mas acho que podemos falar de AL de forma geral. Realmente esse é um resultado ``negativo'', mas não sei como apresentar com um tom melhor}

Thus, RoBERTa outperforms BERT in AL scenarios with low availability of labeled data. However, perhaps more importantly, the experiments demonstrate the superiority of our DoTCAL fine-tuning process w.r.t the traditional 1-step fine-tuning, also when applied to a robust language model like  RoBERTa.

 %\section{Limitations} %moved back to RW

%Finally, as we  address relatively small amounts of labeled data to train models, a related research area is few-shot learning (FSL) \cite{fasl, gu-etal-2022-ppt}, which exploits small amounts of labeled data, especially for neural-network solutions. Most current FSL strategies focus on prompt engineering approaches, such as soft prompt tuning \cite{gu-etal-2022-ppt}. Answering which paradigm (prompt or fine-tuning) is the most robust and effective requires further investigation, being out of the scope of this work.  In any case, adapting AL strategies, for example, to select which instances to use for prompt engineering is promising and left for future work.\looseness=-1

\section{Discussion: Theoretical and Practical Implications of our Study} \label{sec:discussion}

The research presented in this article holds both practical and theoretical implications for supervised automatic text classification (ATC) and active learning (AL).\looseness=-1

From a theoretical perspective,  we first offer an in-depth exploration of the cold-start active learning scenario, which, to the best of our knowledge, is either completely ignored in previous work or evaluated co-jointly with non-cold-start scenarios, which makes it difficult to distinguish the effects of these two scenarios.\looseness=-1

Secondly, we propose DoTCAL, a novel two-step fine-tuning pipeline for contextual embeddings in the context of cold-start AL. This contribution expands the understanding of how to effectively adapt pre-trained models to specific domains with limited labeled data. By leveraging all available unlabeled data and actively labeled samples, we demonstrate the potential for achieving higher classification effectiveness with reduced labeling effort.\looseness=-1

Furthermore, our comparative evaluation of different text representations, including contextual embeddings (BERT and RoBERTa), bag-of-words (BoW), and Latent Semantic Indexing (LSI), sheds light on their effectiveness under various budget scenarios and AL stages (selection and classification). The study provides valuable insights into the strengths and weaknesses of each representation approach in different contexts, contributing to a better understanding of their applicability in ATC tasks with limited labeled data.\looseness=-1

We address important research questions related to the effectiveness of our proposed DoTCAL pipeline and the impact of text representation on AL outcomes. By answering these questions, we advance the theoretical knowledge on designing and implementing effective AL strategies for ATC tasks, especially in situations where labeled data is scarce.\looseness=-1

From a practical perspective, our study can benefit researchers and practitioners working in the areas of ATC and AL, by offering:\looseness=-1

\begin{itemize}

\item  Enhanced Active Learning Effectiveness:  DoTCAL significantly improves the effectiveness of active learning in the cold-start scenario. By leveraging all available unlabeled data and actively labeled samples, our approach reduces the labeling effort while achieving higher classification effectiveness. This proposal has clear practical implications in scenarios and application where labeling large amounts of data can be time-consuming and expensive.\looseness=-1
%Improved AL Effectiveness: The proposed two-step fine-tuning pipeline provides a practical solution for improving AL effectiveness in cold-start scenarios. By employing this approach, researchers and practitioners can reduce the labeling effort required to achieve high classification accuracy, making the annotation process more cost-effective and efficient.

\item Guided Text Representation Choices: Our comparative evaluation of different text representation paradigms offers practical guidance on which approach to use for AL in specific contexts. Depending on the dataset and budget constraints, practitioners can choose between BoW, LSI, or contextual embeddings (BERT or RoBERTa) to optimize AL performance.\looseness=-1

\item Adaptability to Different Domains: The proposed domain-oriented adaptation phase of the two-step fine-tuning pipeline allows the AL process to be tailored to different domains. This adaptability is crucial when labeled data is scarce, and contextual embeddings must be fine-tuned effectively for domain-specific tasks. A possible alternative direction for the AL line of work is the prompt-learning paradigm, particularly when exploiting LLMs, which mainly does not require fine-tuning. However, there is still no consensus on the best paradigm, while  LLMs have limitations regarding privacy issues and higher time and infrastructure costs.\looseness=-1

\item Generalization to other BERT-style language models:  our evaluation demonstrates the superiority of our fine-tuning pipeline over the traditional one-step fine-tuning with BERT-based contextual embeddings and the larger and more robust RoBERTa model. This finding suggests that the benefits of our approach extend beyond specific embedding architectures and can be applied to other large-scale pre-trained language models.\looseness=-1

\end{itemize}

\section{Conclusions and Future Work} \label{sec:conc}

We tackled the ``cold-start'' AL problem for ATC under the (text) representation perspective, aiming
at investigating how different representation approaches (e.g., contextual embeddings, BoW, LSI, FastText) impact AL effectiveness considering different labeling budget sizes, and each AL stage (selection and classification). We proposed DoTCAL, a novel two-step fine-tuning pipeline for cold-start AL which first exploits domain adaptation with unlabeled data, and further adapts the model to the ATC task exploiting the data that was actively labeled after the AL selection stage. 
Our experiments, using eight ATC benchmark datasets, show that DoTCAL outperforms prior pipelines in classification effectiveness by up to 33\%, requiring a much lower labeling effort (typically halving the original labeling costs). We also show that traditional text representations such as BoW and LSI can be useful and effective as input to both selection and classification stages, especially with low budgets and hard-to-classify tasks. We show the benefits of LSI when training with a reduced amount of labeled data, despite the potential loss of information. Finally, we demonstrated that our solutions generalize to other Transformer-Based Language Models such as RoBERTa.

As a future work, we intend to produce novel robust representations that  combine the individual strengths of each approach or automatically select the best representation for each stage based on dataset characteristics with AutoML. We will also perform tests with other Transformers (XLNET, GPT-3). We also intend to perform a cost-benefit analysis on LLMs (effectiveness vs cost) to potentially pursue AL for fine-tuning LLMs. Finally, we intend to apply our proposal in other domains with data scarcity, such as medical~\cite{zanotto2021stroke} and supervised short text topic modeling~\cite{tm1,tm2,tm3}.\looseness=-1

\begin{acks}
This work was partially supported by CNPq, CAPES, FAPEMIG, Amazon Web Services, NVIDIA, CIIA-Saúde, and FAPESP. 
All authors approved the final version of the manuscript.
\end{acks}

\bibliographystyle{ACM-Reference-Format}
\bibliography{main}

\newpage

\section*{Appendix A - Parameter Tuning}

Using the validation sets,  we searched for the parameter values that led to the best macro-F1 results. For the SVM classifier, we set the linear kernel and varied the \textit{C} parameter in \{0.01, 0.1, 1, 10\}. For all datasets and AL configurations, the best choice was $C$=$1$.\looseness=-1

 For the DWDS AL strategy, we varied the distance threshold $\mathit{dist_{min}}$ in \{0.001, 0.01, 0.1, 0.5, 0.6, 0.7, 0.8, 0.9\}. For FastText and BERT-based vectors, the best choice was $\mathit{dist_{min}}=0.01$, while for BoW and LSI-based representations, the best parameter value was $\mathit{dist_{min}}=0.7$.\looseness=-1

For the fine-tuning pipeline, we set the learning rate for both steps $\lambda_{\mathit{MLM}}$=$\lambda_{\mathit{ATC}}$ = $5\times10^{-5}$. We tested with two values for the number of epochs in the domain adaptation step: $e_{\mathit{MLM}}$=$\{10, 20\}$, opting for $e_{\mathit{MLM}}$=$10$, since it provides the same effectiveness with lower cost. Following reference values in previous work~\citep{Cunha_2021,cunha24}, we set the number of epochs $e_{\mathit{ATC}}$=$5$, for both traditional fine-tuning pipeline and the second step of our fine-tuning approach.\looseness=-1

 For LSI-based representations, we set the number of latent dimensions as $d$=$768$, using the same number of dimensions as the BERT-based vectors. Following, we present results for other numbers of latent dimensions.\looseness=-1

\section*{Appendix B - The impact of Latent Dimensions in AL Selection} \label{sec:lsi}

\begin{figure}[!ht]
     \centering
     \begin{subfigure}[t]{0.25\textwidth}
         \centering
         \includegraphics[width=\textwidth]{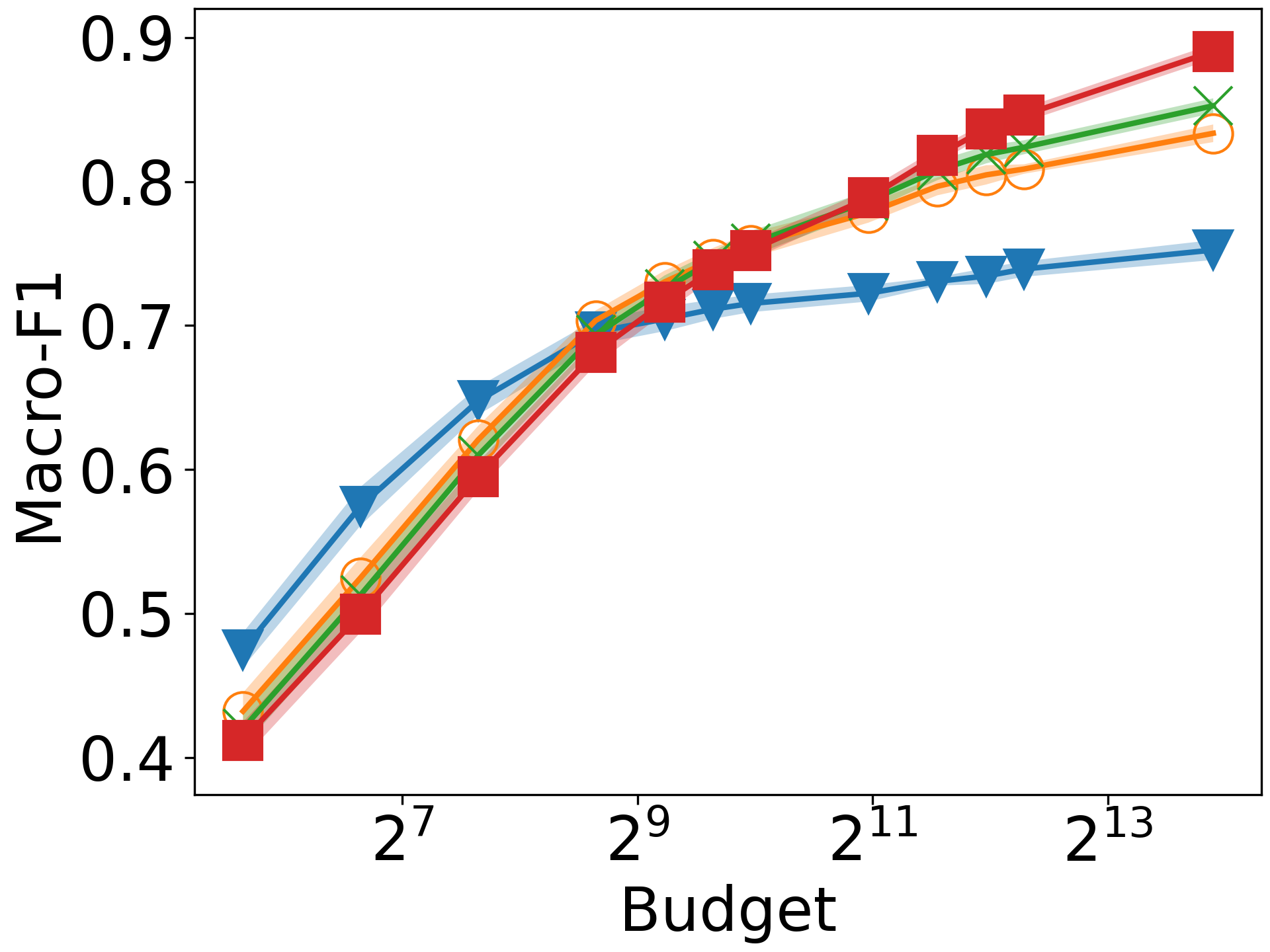}
         \caption{20NG}
         \label{a}
     \end{subfigure}
\hfill
     \begin{subfigure}[t]{0.25\textwidth}
         \centering
         \includegraphics[width=\textwidth]{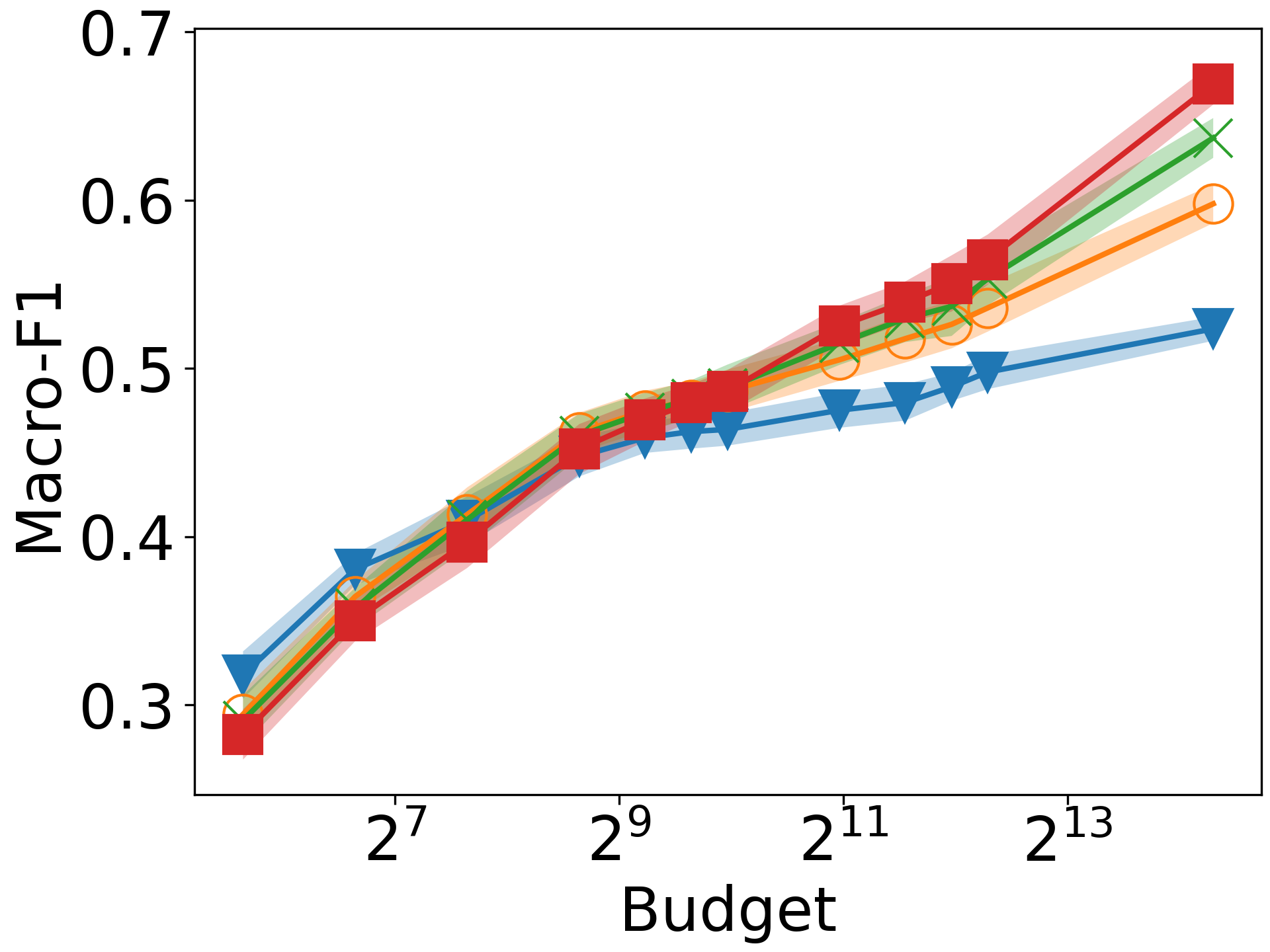}
         \caption{ACM}
         \label{a}
     \end{subfigure}
\hfill
     \begin{subfigure}[t]{0.25\textwidth}
         \centering
         \includegraphics[width=\textwidth]{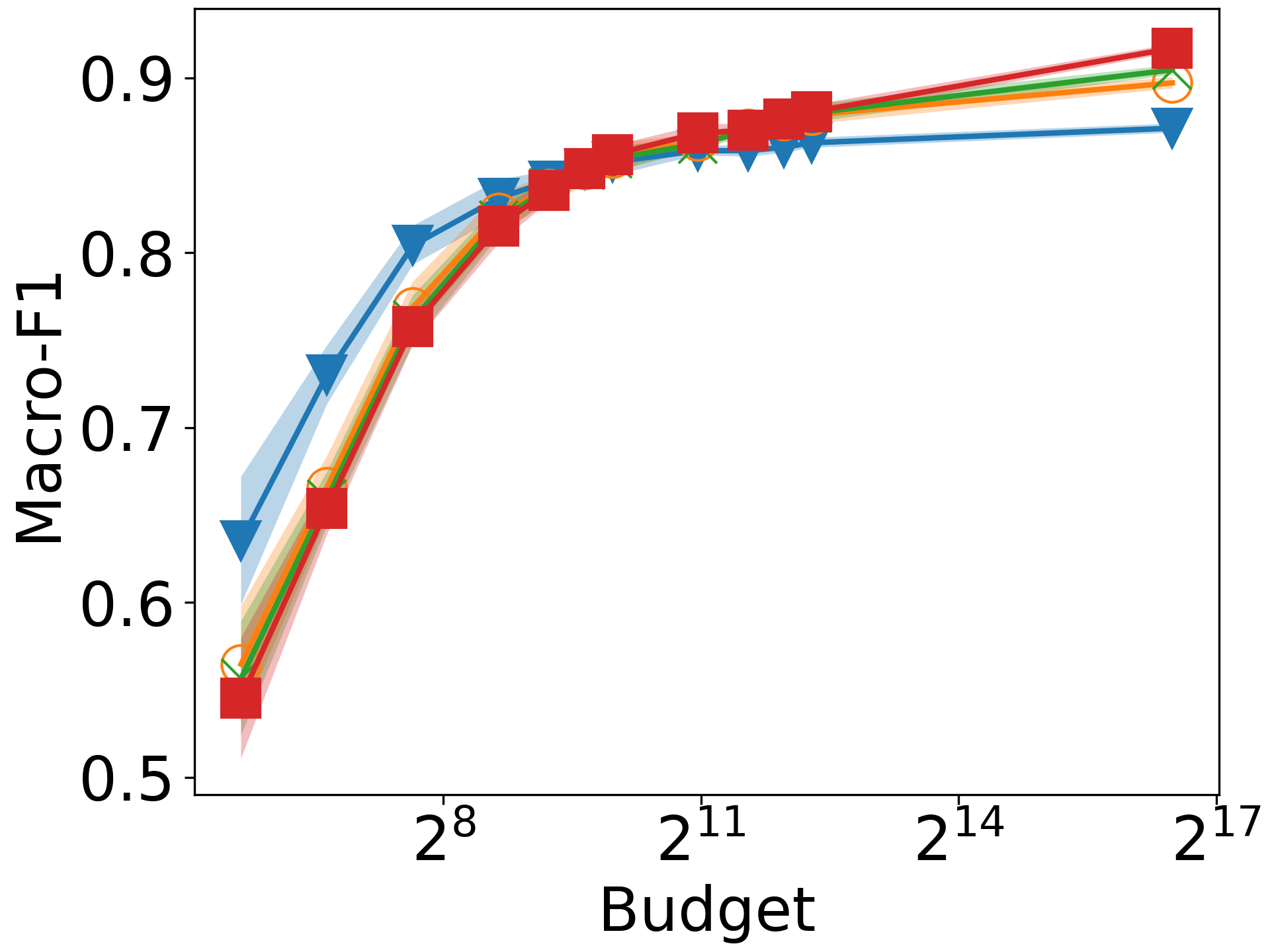}
         \caption{AGNews}
         \label{a}
     \end{subfigure}
\hfill
     \begin{subfigure}[t]{0.25\textwidth}
         \centering
         \includegraphics[width=\textwidth]{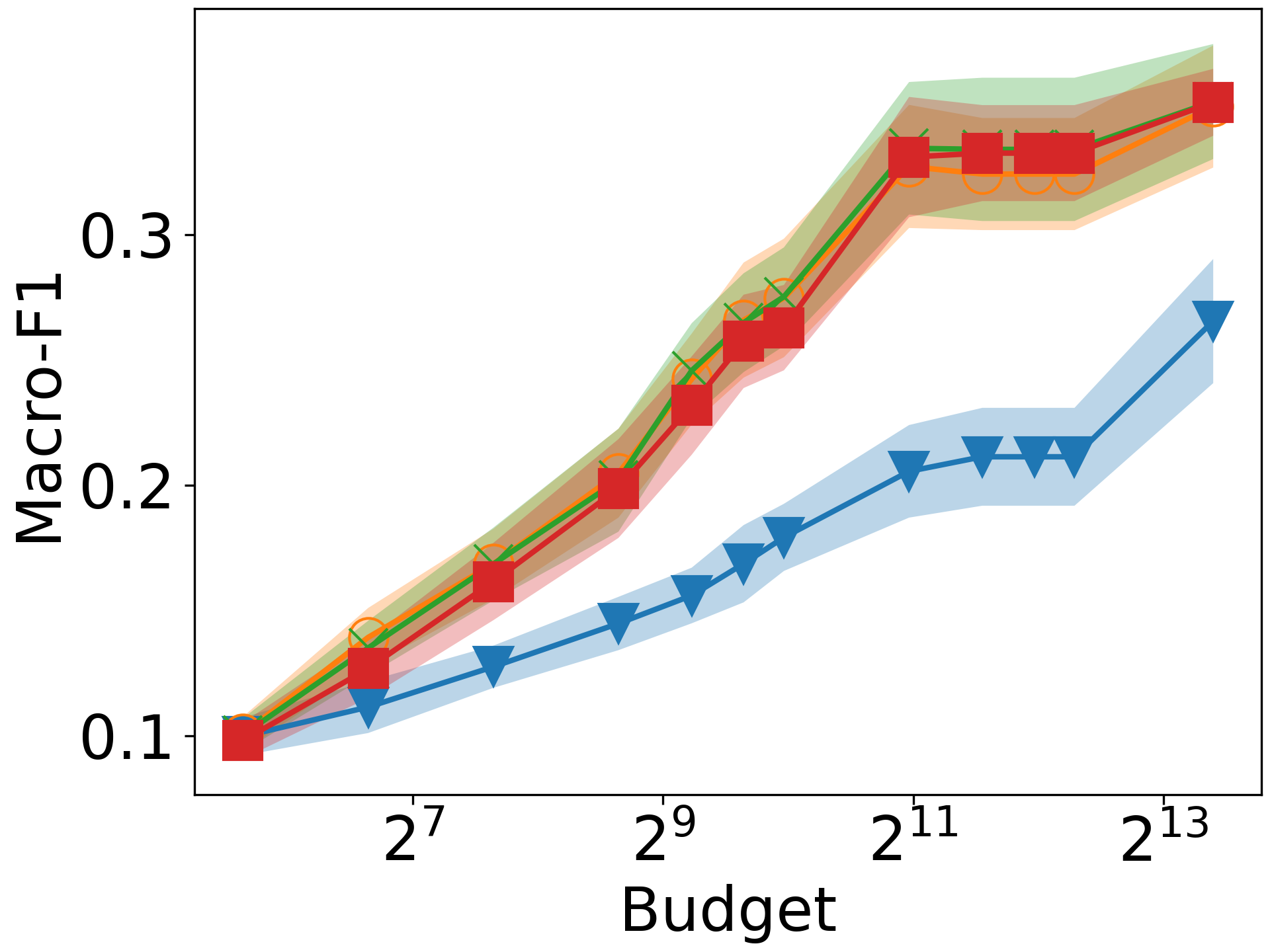}
         \caption{Reuters}
         \label{a}
     \end{subfigure}
%\hfill
%     \begin{subfigure}[t]{0.32\textwidth}
%         \centering
%         \includegraphics[width=\textwidth]{img/lsi_dim/webkb_curves_log2.png}
%         \caption{WebKB}
%         \label{a}
%     \end{subfigure}
%\hfill
%     \begin{subfigure}[t]{0.32\textwidth}
%         \centering
%         \includegraphics[width=\textwidth]{img/lsi_dim/imdb_reviews_curves_log2.png}
%         \caption{IMDB Reviews}
%         \label{a}
%     \end{subfigure}
%\hfill
%     \begin{subfigure}[t]{0.32\textwidth}
%         \centering
%         \includegraphics[width=\textwidth]{img/lsi_dim/vader_movie_2L_curves_log2.png}
%         \caption{VaderMovie}
%         \label{a}
%     \end{subfigure}
\hfill
     \begin{subfigure}[t]{0.3\textwidth}
         \centering
         \includegraphics[width=\textwidth]{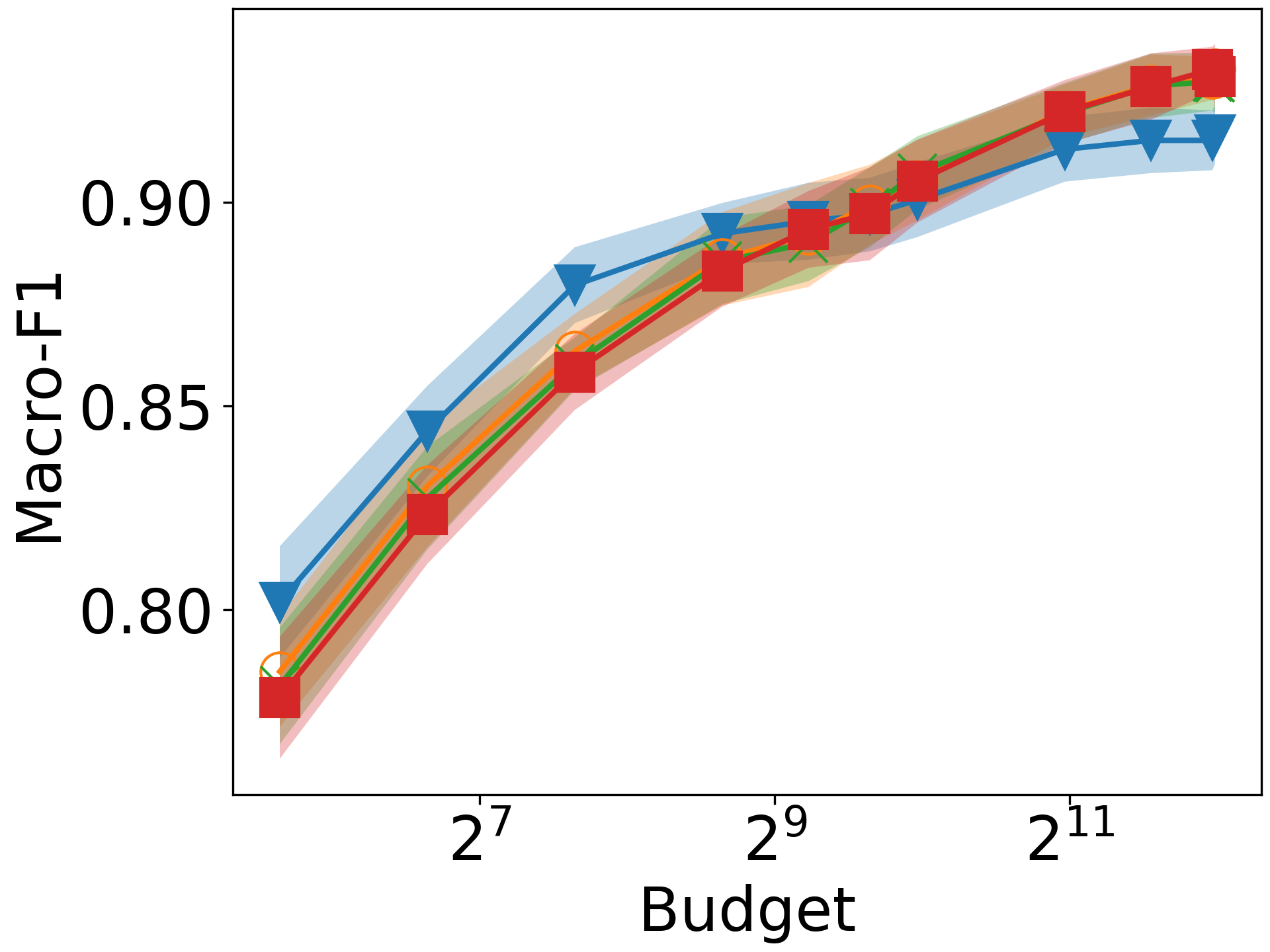}
         \caption{YelpReviews}
         \label{a}
     \end{subfigure}
%\hfill
     \begin{subfigure}[t]{0.15\textwidth}
         \centering
         \includegraphics[width=\textwidth]{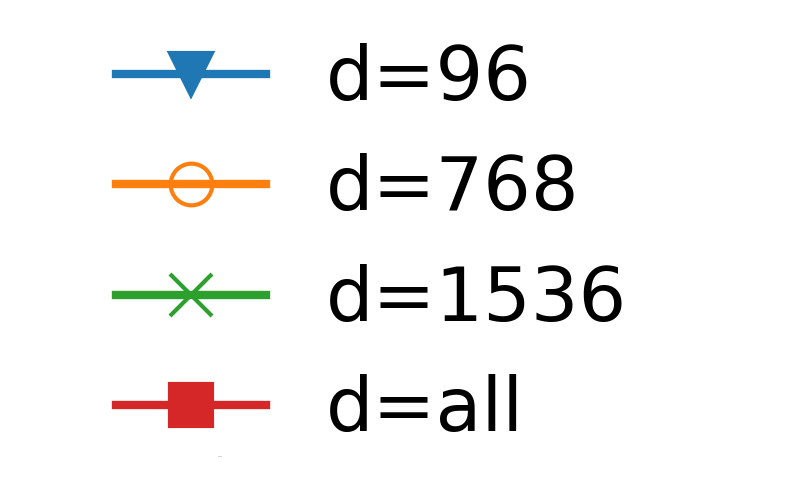}
         %\caption{YelpReviews}
         %\label{a}
     \end{subfigure}
%\hfill
%      \begin{subfigure}[t]{0.31\textwidth}
%          \centering
%          \includegraphics[width=\textwidth]{img/lsi_dim/legend_svd_dim.png}
%          %\caption{YelpReviews}
%          %\label{a}
%      \end{subfigure}

\caption{Macro-F1 for different budgets and different number $d$ of dimensions for the LSI representation. $d$=all means the vocabulary size of each dataset (i.e., the original BoW without compression). 95\% confidence intervals are shown in shaded areas.}
    \label{fig:lsi}

\end{figure}

Figure \ref{fig:lsi} shows average macro-F1 results for different budgets and different numbers of latent dimensions ($d$). For results in this figure, the representation approach employed to select instances was fixed in BoW. Similar results were obtained for the other selection approaches.\looseness=-1

In our experiments, we varied parameter $d$ in \{96, 192, 384, 768, 1536, 3072\}, which comprise various divisors and multipliers of 768, the number of dimensions of the employed BERT-based representations. Figure \ref{fig:lsi} shows results for $d$ in \{96, 768, 1536\}, for the sake of better visualization, and for $d$=all, which corresponds to the original BoW representation without compression.\looseness=-1 

According to our results, the classification effectiveness does not decay when reducing the number of dimensions in the LSI-based representation for scenarios with a reduced amount of labeled data (e.g., under 300 instances). On the contrary, for most datasets, we observe gains of up to 17\% in macro-F1 w.r.t the original, non-compressed vocabulary when using only 96 latent dimensions for small budgets (50 up to 400 instances). This occurs because, in the dimensional reduction process, words that frequently co-occur in the training dataset are combined as ``more compressed'' latent terms, allowing to aggregate information of many instances of the given dataset in a few dimensions, which are potentially common to many instances. This approach is suitable in an AL task, in which we are interested in selecting only a few representative examples from the whole dataset to label. \textit{Compression seems to help emphasizing what is important (or not) for the sake of selecting Representative and diverse instances.}\looseness=-1

Thus, in low-budget AL scenarios, unlike scenarios with many labeled data, it is interesting to exploit LSI using only a few (e.g., 50-100) latent dimensions, which produce gains in comparison with full-dimensional BoW representations, regardless of the potential information loss inherent to this compression process.\looseness=-1

\end{document}